\documentclass[10pt]{article} 
\usepackage[accepted]{tmlr}

\usepackage{amsmath,amsfonts,bm}

\def\eqref#1{equation~\ref{#1}}

\def\1{\bm{1}}

\DeclareMathAlphabet{\mathsfit}{\encodingdefault}{\sfdefault}{m}{sl}
\SetMathAlphabet{\mathsfit}{bold}{\encodingdefault}{\sfdefault}{bx}{n}

\usepackage{hyperref}
\usepackage{url}

\usepackage{microtype}
\usepackage{graphicx}
\usepackage{subfig}
\usepackage{booktabs} 
\usepackage[font=small,skip=-1.pt]{caption}
\usepackage{wrapfig}
\usepackage{multirow}
\usepackage{amsmath}
\usepackage{lastpage}
\usepackage[ruled,vlined]{algorithm2e}
\usepackage{bm}
\usepackage{multirow}

\usepackage{xcolor}
\newcommand{\rev}[1]{{#1}}

\makeatletter
\DeclareRobustCommand\onedot{\futurelet\@let@token\@onedot}
\def\@onedot{\ifx\@let@token.\else.\null\fi\xspace}

\makeatother

\usepackage[capitalize]{cleveref}
\crefname{section}{Sec.}{Secs.}
\Crefname{section}{Section}{Sections}
\Crefname{table}{Table}{Tables}
\crefname{table}{Tab.}{Tabs.}

\newcommand{\methodname}{CacheFlow}

\title{CacheFlow: Fast Human Motion Prediction by Cached Normalizing Flow}

\author{\name Takahiro Maeda$^1$ \thanks{Currently at Toshiba Corporation.} \email takahiro.maeda.k91@mail.toshiba \\
\name Jinkun Cao$^2$ \email jinkunc@andrew.cmu.edu \\
\name Norimichi Ukita$^1$ \email ukita@toyota-ti.ac.jp \\
\name Kris Kitani$^2$ \email kmkitani@andrew.cmu.edu \\
\addr $^1$ Toyota Technological Institute \\
\addr $^2$ Robotics Institute, Carnegie Mellon University 
}

\def\month{01}  
\def\year{2026} 
\def\openreview{\url{https://openreview.net/forum?id=y8dGdBJkWa}} 

\begin{document}

\maketitle

\begin{abstract}
Many density estimation techniques for 3D human motion prediction require a significant amount of inference time, often exceeding the duration of the predicted time horizon. 
To address the need for faster density estimation for 3D human motion prediction, we introduce a novel flow-based method for human motion prediction called CacheFlow.
Unlike previous conditional generative models that suffer from poor time efficiency, CacheFlow takes advantage of an unconditional flow-based generative model that transforms a Gaussian mixture into the density of future motions.
The results of the computation of the flow-based generative model can be precomputed and cached.
Then, for conditional prediction, we seek a mapping from historical trajectories to samples in the Gaussian mixture.
This mapping can be done by a much more lightweight model, thus saving significant computation overhead compared to a typical conditional flow model.
In such a two-stage fashion and by caching results from the slow flow model computation, we build our CacheFlow without loss of prediction accuracy and model expressiveness. 
This inference process is completed in approximately one millisecond, making it 4$\times$ faster than previous VAE methods and 30$\times$ faster than previous diffusion-based methods on standard benchmarks such as Human3.6M and AMASS datasets.
Furthermore, our method demonstrates improved density estimation accuracy and comparable prediction accuracy to a SOTA method on Human3.6M.
Our code and models are available at \url{https://github.com/meaten/CacheFlow}.
\end{abstract}

\section{Introduction}
\label{sec:intro}

The task of 3D human motion prediction is to forecast the future 3D pose sequence given an observed past sequence.
Traditional motion prediction methods are often based on deterministic models and can struggle to capture the inherent uncertainty in human movement. 
Recently, stochastic approaches have addressed this limitation. Stochastic approaches allow models to sample multiple possible future motions.
Stochastic human motion prediction methods utilize conditional generative models such as generative adversarial networks (GANs)~\citep{gan}, variational autoencoders (VAEs)~\citep{vae}, and denoising diffusion probabilistic model~\citep{ddpm}.
However, many stochastic approaches cannot explicitly model the probability density distribution.

Conversely, density estimate-based approaches explicitly model the probability density distribution.
In safety-critical applications such as autonomous driving~\citep{survey_autonomous_driving} and human-robot interaction~\citep{koppula2015anticipating, lasota2017multiple, butepage2018anticipating}, a density estimate can represent all possible future motions (not just a few samples) by tracking the volume of density. It can be used to derive guarantees on safety~\citep{mainprice2013human, sanderud2015likelihood, tisnikar2024probabilistic}.

However, previous density estimation suffers from high computational cost. The expensive computational cost can prohibit applications to real-time use-cases, especially with high dimensional data such as human motions.
For instance, kernel density estimation (KDE)~\citep{rosenblatt1956remarks, parzen1962estimation} requires an exponentially growing number of samples for accurate estimation.
Concretely, more than one trillion samples are required for accurate KDE over a 48-dim pose over 100 frames of human motion prediction~\citep{silverman2018density}.

In contrast to traditional KDE, recent parametric density estimation approaches use conditional flow-based generative models, including normalizing flows~\citep{rezende2015variational, tabak2013family, tabak2010density} and continuous normalizing flows~\citep{chen2018neural}.
These flow-based generative models (``flow-based model'' for brevity) directly estimate the density to avoid time-consuming sampling required in KDE.
However, inferring the exact probability of possible future motions remains computationally expensive.
This is because capturing the full shape of the distribution requires evaluating the probabilities of many potential future motions.

\begin{figure}[t]
    \centering
    \includegraphics[width=\textwidth]{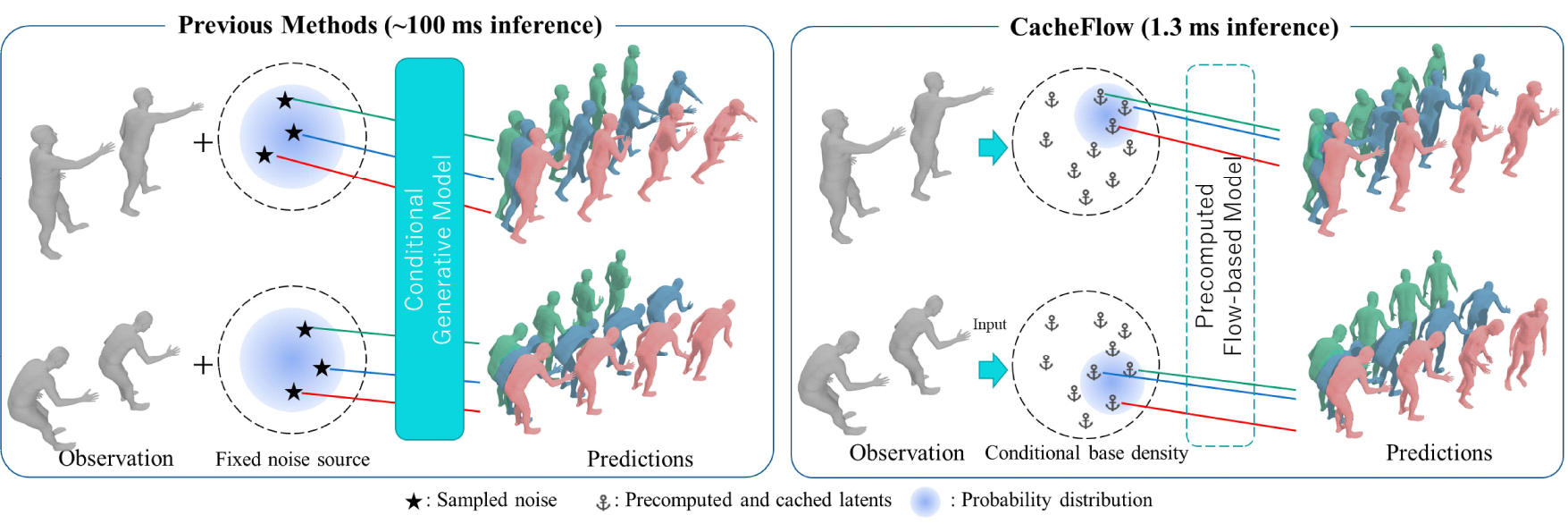}
    \caption{\textbf{Previous methods vs. Our \methodname{}.}
    Previous methods of stochastic motion prediction generate multiple future motions by sampling noises from the fixed source in an ad hoc manner. In contrast, \methodname{} uses the precomputed and cached latent-motion pairs from an unconditional flow-based generative model.
    Thus, the computation of the unconditional flow can be skipped at inference.
    One can achieve fast inference by selecting predictions from these cached pairs.}
    \label{fig:teaser_real}
\end{figure}

To address this computational limitation, we propose a fast density estimation method based on a flow-based model called "\methodname{}".
Our \methodname{} utilizes an unconditional flow-based model for prediction, as illustrated in \cref{fig:teaser_real}.
Since the unconditional flow-based model is independent of past observed motions, its calculation can be precomputed and skipped at inference.
This precomputation omits a large portion of computational cost.
To achieve further acceleration, our unconditional flow-based model represents transformation between a lightweight conditional base density and the density of future motions.
At inference, the density of future motion is estimated by computing the lightweight conditional base density and combining it with the precomputed results of the flow-based model.
The inference of our method is approximately one millisecond.

\methodname{} demonstrates comparable accuracy to previous methods on standard stochastic human motion prediction benchmarks, Human3.6M~\citep{ionescu2013h36m} and AMASS~\citep{mahmood2019amass}.
Furthermore, our method estimates density more accurately than previous stochastic human motion prediction methods with KDE.
\methodname{} shows improved computational efficiency, making it well-suited for real-time applications.
The contributions of this paper are fourfold as follows:

\begin{enumerate}
    \item We introduce a novel fast density estimation called \methodname{} on human motion prediction.
    \item We can sample diverse future motion trajectories with explicit density estimation, and we experimentally confirm that our method can estimate accurate density.
    \item Our method achieves comparable prediction accuracy to other computationally intense methods on several benchmarks. 
\end{enumerate}

\section{Related Work}
\label{sec:related_work}
\subsection{Human Motion Prediction}
\textbf{Deterministic approaches.}
Early approaches on human motion prediction~\citep{aksan2021spatio,bouazizi2022motionmixer,fragkiadaki2015recurrent,butepage2017deep,guo2023back,jain2016structural,li2018convolutional} focused on deterministic settings.
They predict the most likely motion sequence based on the past motion.
A wide range of architectures were proposed including multi-layer perceptron~\citep{guo2023back}, recurrent neural networks~\citep{fragkiadaki2015recurrent,jain2016structural,martinez2017human,gui2018adversarial,pavllo2018quaternet,liu2019towards}, convolutional neural networks~\citep{li2018convolutional,medjaouri2022hr}, transformers~\citep{aksan2021spatio,cai2020learning,martinez2021pose}, and graph neural networks (GNNs)~\citep{mao2019learning,li2020dynamic,dang2021msr,li2021skeleton}.
GNN can account for the explicit tree expression of the human skeleton, while other architectures implicitly learn the dependencies between joints.

\noindent \textbf{Stochastic approaches.}
To capture the inherent uncertainty in human movements, recent works have focused on stochastic human motion prediction to predict multiple likely future motions.
The main stream of stochastic methods use generative models for the purpose, such as generative adversarial networks (GANs)~\citep{barsoum2018hp, kundu2019bihmp}, variational autoencoder (VAE)~\citep{walker2017pose, yan2018mt, mao2021generating, cai2021unified}, and denoising diffusion probabilistic model (DDPM)~\citep{barquero2023belfusion,chen2023humanmac,wei2023human,sun2024comusion}.
To improve the diversity of predictions, diversity-promoting loss~\citep{mao2021generating,barquero2023belfusion} or explicit sampling techniques~\citep{wei2023human} were proposed.
In contrast to generative models, anchor-based methods~\citep{anchor,xu2024learning} learn a fixed number of anchors corresponding to each prediction to ensure diversity.
However, most stochastic methods cannot describe the density of future motions explicitly.
This prevents exhaustive or maximum likelihood sampling for practical applications.
On the contrary, our method allows for explicit density estimation using normalizing flows~\citep{kobyzev2020normalizing}.

\subsection{Density Estimation}

Density estimation asks for explicit calculation of the probability for samples from a distribution.
Density estimation is derived by non-parametric or parametric methods.

\noindent  \textbf{Non-parametric Approach.}
The representative non-parametric density estimation is kernel density estimation (KDE)~\citep{rosenblatt1956remarks, parzen1962estimation}.
KDE can estimate density by using samples from generative models.
However, KDE requires a large number of samples for accurate estimation.
Therefore, it often cannot run in real-time.

\noindent \textbf{Parametric Approach.}
As a representative parametric model, Gaussian mixture models (GMMs) parametrize density with several Gaussian distributions and their mixture weights.
Its nature of mixing Gaussian priors limits its ability to generalize to complex data distribution. 
Another parametric approach with more expressivity is flow-based generative models~\citep{kobyzev2020normalizing}.
By a learned bijective process, normalizing flows (NFs)~\citep{rezende2015variational, tabak2013family, tabak2010density} transform a simple density like the standard normal distribution into a complex data density.
Recently, continuous normalizing flows (CNFs)~\citep{chen2018neural,DBLP:conf/iclr/GrathwohlCBSD19} achieve more expressive density than standard normalizing flows via an ODE-based bijective process.
While training of CNFs is inefficient due to the optimization of ODE solutions, an efficient training strategy named flow matching~\citep{DBLP:conf/iclr/LipmanCBNL23} was proposed.
FlowChain~\citep{maeda2023fast} was proposed for fast and efficient density estimation in human trajectory forecasting.
FlowChain improves the inference time efficiency by reusing results from the conditional flow-based method while the past sequences are similar. 
However, with significantly different past sequences, FlowChain's efficiency can't hold anymore.
Unlike FlowChain, our method can perform fast and efficient inference regardless of past sequences. 



\subsection{\rev{Diffusion-based Motion Modeling}}

Recently, diffusion-based generative models have achieved state-of-the-art performance in synthesizing diverse and high-fidelity human motion. The Motion Diffusion Model (MDM)~\citep{mdm} demonstrated that diffusion processes can model complex kinematic distinctness, capable of tasks ranging from text-to-motion to motion completion. Concurrently, approaches inspired by Large Language Models, such as MotionGPT~\citep{motiongpt} and MotionLM~\citep{motionlm}, treat motion sequences as discrete tokens, leveraging the transformer architecture to capture long-term semantic dependencies.

While these methods excel in generation quality and semantic control, they often incur high computational costs during inference. Diffusion models typically require multi-step iterative denoising, and LLM-based approaches rely on heavy autoregressive sampling. In contrast to these heavy generative frameworks, our work focuses specifically on the efficiency-latency trade-off. While we share the goal of modeling complex motion distributions, CacheFlow leverages the tractable likelihoods of Normalizing Flows and introduces caching mechanisms to ensure rapid inference suitable for real-time prediction scenarios, avoiding the computational bottleneck of iterative diffusion or large-scale transformers.

\section{Preliminary}
\label{sec:preliminary}

\subsection{Problem Formulation}

The task of human motion prediction aims to use a short sequence of observed human motion to predict the future unobserved motion sequence of that person.
Human motion is represented by a sequence of human poses in a pre-defined skeleton format of 3D locations of $J$ joints, $X \in \mathbb{R}^{J \times 3}$.
As input to our model, we have the past (history of) human motion as a sequence $\bm{c}=[X_{1}, X_{2}, ...,X_H]\in\mathbb{R}^{H \times J \times 3}$ over $H$ timesteps.
To predict the future human motion sequence of $F$ timesteps, we can formulate the problem as one of conditional generation using the conditional probability function, $p(\bm{X}|\bm{c})$, where $\bm{X}=[X_{H+1}, X_{H+2}, ...,X_{H+F}]\in\mathbb{R}^{F \times J \times 3}$. 
Similar to the stochastic human motion prediction paradigm, the method should also allow for sampling $n$ multiple future sequences $\{\bm{X}_1, ..., \bm{X}_n\}$ from $p(\bm{X}|\bm{c})$.
The focus of our work is to accelerate the inference time of estimate and sampling of the conditional density function $p(\bm{X}|\bm{c})$.



%
\subsection{Normalizing Flow}
\label{subsec:normalizing_flow}
Normalizing flow~\citep{rezende2015variational, tabak2013family, tabak2010density} is a generative model with explicit density estimation. 
It follows a bijective mapping \(f_\theta\) with learnable parameters $\theta$.
It transforms a simple base density \(q(\bm{z})\) such as a Gaussian distribution into the complex data density \(p(\bm{x})\).
We can analytically estimate the exact probability via the change-of-variables formula as follows:
\begin{align}
    \bm{x} &= f_\theta(\bm{z}), \quad \bm{z} =f_\theta^{-1}(\bm{x})\label{eq:tranform}.\\
    p(\bm{x}) &= q(\bm{z})|{\rm det}\mathcal{J}_{f_\theta}(\bm{z})|^{-1} \label{eq:prob},
\end{align}
where $\mathcal{J}_{f_\theta}(z) = \frac{\partial f_\theta}{\partial \bm{z}}$ is the Jacobian of $f_\theta$ at $\bm{z}$.
The parameters $\theta$ of \(f_\theta\) can be learned by maximizing the likelihood (or conditional likelihood) of samples \(\hat{\bm{x}}\) from datasets or minimizing the negative log-likelihood as $\mathcal{L}_\text{NLL} = - \log p(\hat{\bm{x}})$.
When $\bm{x}$ and $\bm{z}$ are latent codes, normalizing flow is transformed into latent normalizing flow. We follow this pattern in our method. We encode the past human motion into $\bm{x}$ by an encoder network $\mathcal{E}$ and decode it by a decoder network $\mathcal{D}$:

\begin{equation}
    \bm{x} = \mathcal{E}(\bm{X}), \bm{X} =  \mathcal{D}(\bm{x}). \quad \bm{x} \sim \mathbb{R}^{d}, \bm{X} \sim \mathbb{R}^{F \times J \times 3}
\end{equation}
The encoder and decoder are trained by reconstruction. In the later part of this paper, for simplicity, we discuss the method at the latent representation level and model the conditional generation task as $p(\bm{x}|\bm{c})$.

\subsection{Continuous Normalizing Flow (CNF)}
\label{subsec:cnf}
Continuous normalizing flow~(CNF)~\citep{chen2018neural,DBLP:conf/iclr/GrathwohlCBSD19} is a normalizing flow variant based on an ordinary differential equation~(ODE).
CNF defines $t$-continuous path $\bm{z}_t$ between the base density space $\bm{z}_0 \sim q(\bm{z})$ and the data space $\bm{z}_1 = \bm{x} \sim p(\bm{x})$.
This $\bm{z}_t$ is defined by the parameterized vector field $\frac{d \bm{z}_t}{dt} = v_\theta(\bm{z}_t)$.
The data $\bm{x}=\bm{z}_1$ is generated via numerical integration of vector field $v_\theta(\bm{z}_t)$ as follows:
\begin{align}
    \bm{x} = \bm{z}_1 = \bm{z}_0 + \int_0^1 v_\theta(\bm{z}_t) dt \label{eq:cnf}.
\end{align}
The CNF transformation \cref{eq:cnf} is denoted as $\bm{x}=f_\theta(\bm{z})$ for brevity.
Although CNF can be trained by minimizing negative log-likelihood, it is time-consuming due to the numerical integration of ODE.

\begin{figure}[t]
    \centering
    \includegraphics[width=\textwidth]{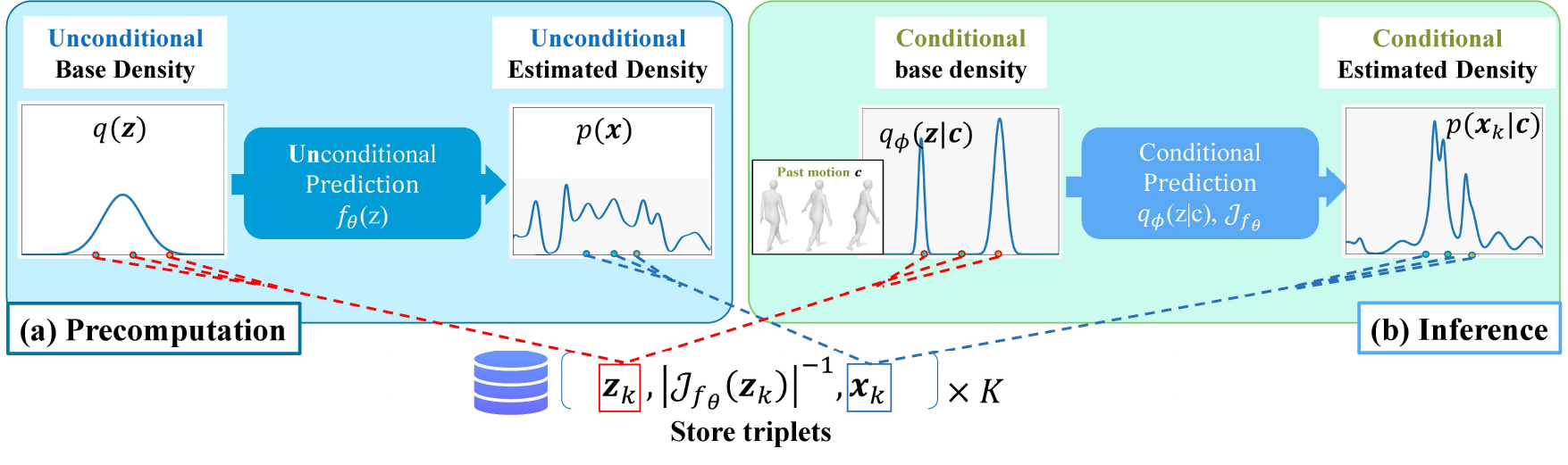}
    \caption{\textbf{Overview of our \methodname{}.}
    Our method utilizes the unconditional flow-based model $f_\theta$.
    This $f_\theta$ maps the lightweight conditional base density $q_\phi(\bm{z}|\bm{c})$ into future motion density $p(\bm{x}|\bm{c})$.
    In this formulation, the flow-based model is independent of past motions.
    Thus, we can precompute the unconditional flow-based model.
    These results are cached as K triplets as shown in (a).
    Due to the precomputation, we can skip the inference of $f_\theta$ and omit a large portion of the entire computation.
    At inference, density estimation is achieved by only evaluating the lightweight conditional base density $q_\phi(\bm{z}_k|\bm{c})$ and combining it with the stored K triplets as shown in (b).}
    \label{fig:overview}
\end{figure}

\subsection{Flow Matching}
\label{subsec:flowmatching}
In order to train the parameterized vector field efficiently, one can leverage the flow matching~\citep{DBLP:conf/iclr/LipmanCBNL23} strategy.
As a new training strategy, Flow Matching avoids the numerical integration of ODE by directly optimizing the vector field $v_\theta(\bm{z}_t)$.
The objective of flow matching is to match the parameterized vector field $v_\theta(\bm{z}_t)$ to the ground truth vector field $u(\bm{z}_t)$ via mean squared error as follows:
\begin{align}
    \mathcal{L}_\text{FM} = \mathbb{E}_{t \sim \mathcal{T}(0, 1), \bm{z}_t} || v_\theta(\bm{z}_t) - u(\bm{z}_t)||^2 \label{eq:flow_matching},
\end{align}
where $\mathcal{T}(0, 1)$ is a distribution ranging from 0 to 1, such as a uniform distribution.

However, we cannot obtain
the ground truth vector field $u(\bm{z}_t)$ directly.
\citet{DBLP:conf/iclr/LipmanCBNL23} suggest defining the conditional ground truth $u(\bm{z}_t|\hat{\bm{z}}_1)$ instead.
Specifically, it is modeled as a straight vector field $\hat{\bm{z}}_1 - \bm{z}_0$ in \rev{Rectified Flow~\citep{DBLP:conf/iclr/LiuG023}}.
This is trained by the following objective:
\begin{equation}
    \mathcal{L}_\text{CFM} = \mathbb{E}_{t\sim \mathcal{T}(0, 1), \hat{\bm{z}}_1, \bm{z}_t} || v_\theta(\bm{z}_t) - u(\bm{z}_t|\hat{\bm{z}}_1)||^2
    \label{eq:conditional_flow_matching}.
\end{equation}
The gradients of
$\mathcal{L}_\text{FM}$ of \cref{eq:flow_matching} and
$\mathcal{L}_\text{CFM}$ of \cref{eq:conditional_flow_matching} are identical \textit{w.r.t} $\bm{\theta}$.
We exploit expressive CNFs with efficient Flow Matching training to estimate the future motion density $p(\bm{x}|\bm{c})$.
%

\section{Proposed Method}
\label{sec:proposed_method}
%

%


\subsection{Overview of \methodname{}}

\rev{As mentioned in Sec.~\ref{subsec:normalizing_flow}, we propose a flow-based method on a latent representation using frozen $\mathcal{E}$ and $\mathcal{D}$.} We estimate the future motion density $p(\bm{x}|\bm{c})$ by transforming a conditional base distribution $q_\phi(\bm{z}|\bm{c})$, .
This $q_\phi$ is conditioned on the past motion $\bm{c}$.
Then we can sample predictions $\bm{x} \sim p(\bm{x}|\bm{c})$ for stochastic human motion prediction.
Most traditional approaches based on conditional generative models use a trivial source distribution, often a simple Gaussian. However, we redefine the source distribution to be more informative and directly regressed from past motions.
This allows us to develop a much lighter and faster model for predicting future movements.

To build this informative conditional base distribution $q_\phi$, we would incorporate an unconditional flow-based model $f_\theta: \bm{x}=f_\theta(\bm{z})$ that maps latent variable $\bm{z}$ into motion representation $\bm{x}$.
To understand how $q_\phi$ and $f_\theta$ are connected, we first reparametrize the future motion density $p(\bm{x}|\bm{c})$ by a change of variables of probability equation as follows:
\begin{align}
    p(\bm{x}|\bm{c}) 
    &= q(\bm{z}|\bm{c})\bigg|{\rm det}\cfrac{\partial \bm{z}}{\partial \bm{x}}\bigg|,\\
    &=q(\bm{z}|\bm{c})\bigg|{\rm det}\bigg(\cfrac{\partial f_\theta(\bm{z})}{\partial \bm{z}}\bigg)^{-1}\bigg| ,\\
    &= q(\bm{z}|\bm{c}) |{\rm det}\mathcal{J}_{f_\theta}(\bm{z})|^{-1}.
    \label{eq:reparam}
\end{align}
This parametrization trick differs from the widely-used conditional density formulation~\citep{conditional_normalizing_flow} where $\bm{c}$ is conditioned to the flow-based model $f_\theta$.
In this formulation, only the conditional base density $q(\bm{z}|\bm{c})$ varies depending on $\bm{c}$ during inference, whereas the unconditional flow-based model $\bm{x}=f_\theta(\bm{z})$ and the Jacobian $|{\rm det}\mathcal{J}_{f_\theta}(\bm{z})|^{-1}$ are kept same during inference and thus can be reused as-is once calculated.

Therefore, we could precompute the mapping results and Jacobians of an unconditional flow-based model  $f_\theta$.
We cache the triplets $t=\{\bm{z}, |{\rm det}\mathcal{J}_{f_\theta}(z)|^{-1}, \bm{x}\}$ for later reuse in the inference stage, as shown in \cref{fig:overview}(a).

Then, during inference, we design a new trick to reuse the cached triplets by associating them with the specific conditions of the past motion sequences, as shown in \cref{fig:overview}(b).
Now, instead of a typical conditional generative model, e.g., conditional normalizing flow, we only need a lightweight model to model the conditional base density $q_\phi(\bm{z}|\bm{c})$ and achieve similar expressivity. 
We could finally estimate the future motion density by $p(\bm{x}|\bm{c})=q_\phi(\bm{z}|\bm{c})|{\rm det}\mathcal{J}_{f_\theta}(\bm{z})|^{-1}$.
%
The method is summarized as pseudocode in \cref{alg:inference}.
In the following paragraphs, we elaborate on the details of our method. 

\subsection{Precompute Unconditional Flow-based Model}
As the first step of our method, we use the human motion dataset to learn an unconditional flow-based model $f_\theta$. 
From this unconditional human motion prediction model, we will collect the triplets $t=\{\bm{z}, |{\rm det}\mathcal{J}_{f_\theta}(\bm{z})|^{-1}, \bm{x}\}$ for later use. This part is illustrated in \cref{fig:overview}(a).

In our implementation, we built the unconditional flow model by CNFs due to its proven expressivity for predicting human motion.
The unconditional model is trained to predict a fixed-length future motion $\bm{x}$ given a noise sample $\bm{z}$ \rev{sampled from a source distribution $q(\bm{z})$, a standard Gaussian distribution}:
\begin{equation}
    f_\theta: \mathbb{R}^{d} \longrightarrow \mathbb{R}^{d}
\end{equation}

Because $\bm{z}$ is sampled from a known distribution and normalizing-flow models are deterministic with reversible bijective transformation, we could know the density of each $\{\bm{z}, \bm{x}\}$ pair.
We train the unconditional continuous normalizing flow with the flow matching objective described in \cref{eq:conditional_flow_matching}.
Then we collect $K$ samples denoted by the triplet $t_k = \{\bm{z}_k, |{\rm det} \mathcal{J}_{f_\theta}(\bm{z}_k)|^{-1},  \bm{x}_k\}$. 
Triplets are collected by applying the inverse transform of $f_\theta$ to ground truth future motions in the training split.
These triplets are cached for fast inference as described in \cref{subsec:efficient_density}.
This caching operation is different from anchor-based methods~\citep{anchor, xu2024learning} since \methodname{} caches all motions of the training split.

\subsection{Conditional Inference by \methodname{}}
\label{subsec:efficient_density}
In previous methods, conditional human motion prediction typically requires a conditional generative model.
For instance, it is a conditional flow-based or diffusion model.
These models usually have poor time efficiency due to delicate but heavy architecture.
Instead, inspired by \cref{eq:reparam}, we can reuse the results of unconditional inverse transformation as triplets $t_k = \{\bm{z}_k, |{\rm det} \mathcal{J}_{f_\theta}(\bm{z}_k)|^{-1},  \bm{x}_k\}$.
Thus, we can perform conditional inference by only evaluating a conditional base distribution $q(\bm{z}|\bm{c})$.
We model this conditional base distribution by a learnable model, thus we denote it as $q_\phi(\bm{z}|\bm{c})$.
This model can be very lightweight since the unconditional transformation $f_\theta$ gives enough expressivity.
$q_\phi(\bm{z}|\bm{c})$ runs much faster than a typical conditional generative model for human motion prediction.
This part is illustrated in \cref{fig:overview}(b).

In our implementation, $q_\phi(\bm{z}|\bm{c})$ is constructed as a parametrized Gaussian mixture $\{\mathcal{N}(\mu_m(\bm{c}), \sigma^2_m(\bm{c}))\}$, with $M$ mixture weights $w_m(\bm{c})$, such that $\sum_{m=1}^M w_m = 1$.
Each $\mu_m$ and $\sigma_m$ are regressed based on the past motion $\bm{c}$. We use a lightweight single-layer RNN for regression to determine the GMM composition. \rev{This RNN takes the past motion directly as input, and regress the $w_m, \mu_m, \sigma^2_m$. Therefore, $\phi$ means the parameter of this RNN in our method.} Although the unconditional flow-based model $f_\theta$ and the conditional base density $q_\phi$ can be trained separately, we found that jointly training $f_\theta$ and $q_\phi$ improves model performance.
We train the joint model by summation of log-likelihood for $q_\phi$ and flow matching for $f_\theta$ as explained in \cref{eq:conditional_flow_matching} as follows: 
\begin{align}
    \mathcal{L} = -\log q_\phi(f^{-1}_\theta(\hat{\bm{x}})|\bm{c})  + \mathcal{L}_{\rm CFM}.
\end{align}
With joint learning, $f_\theta$ learns an easy mapping for the conditional Gaussian mixture $q_\phi$.

With $q_\phi$ constructed, during inference, we can estimate the conditional density $p(\bm{x}|\bm{c})$ by connecting with precomputed triplets $t_k = \{\bm{z}_k, |{\rm det} \mathcal{J}_{f_\theta}(\bm{z}_k)|^{-1},  \bm{x}_k\}$ as 
\begin{equation}
    p(\bm{x}_k|\bm{c}) = q_\phi(\bm{z}_k|\bm{c})|{\rm det} \mathcal{J}_{f_\theta}(\bm{z}_k)|^{-1}.
\end{equation}
By this inference process, we could optionally generate a future human motion sequence $\bm{x}$ by retrieving a high-probability sample $\bm{z}$ from $q_\phi$ with the past motion sequence as the condition.
However, $q_\phi$ describes a continuous distribution and the stored triplets cannot cover all samples.
Therefore, in practice, predicted motion $\bm{x}_{k^*}$ is selected by the nearest neighbor of the sampling outcome of $q_\phi$ to the stored triplets:
\begin{equation}
\begin{split}
    k^* &= {\rm argmin}_k ||\bm{z}_k - \bm{z}||, \\ 
    {\rm s.t.} \quad \{t_k\ &= \{\bm{z}_k, |{\rm det} \mathcal{J}_{f_\theta}(\bm{z}_k)|^{-1},  \bm{x}_k\}, \quad \bm{z} \sim q_\phi(\bm{z}|\bm{c})\},
\end{split}
\end{equation}
where $k^*$ is the selected index of the triplets for prediction.
By this design, we can sample an arbitrary number of likely future motion sequences by selecting the neighbors of samples $\bm{z} \sim q_\phi(\bm{z}|\bm{c})$.
%
%
%

\begin{algorithm}[t]
    \caption{Precomputation and Inference of \methodname{}.}
    \label{alg:inference}
    \SetAlgoLined
    \DontPrintSemicolon
    \textbf{Input:} Past motion $\bm{c}$\;
    \textbf{Output:} Estimated density $p(\bm{x}_k|\bm{c})$\;
    // Precomputation. This does not count for inference time.

    \For {each future motion $\bm{X}_k$ in the training dataset} {

        $\bm{x}_k \gets \mathcal{E}(\bm{X}_k)$

        $\bm{z}_k \gets f_\theta^{-1}(\bm{x}_k)$

        Calculate $|{\rm det} \mathcal{J}_{f_\theta}(\bm{z}_k)|^{-1}$

        Store triplet $\{\bm{z}_k, |{\rm det} \mathcal{J}_{f_\theta}(\bm{z}_k)|^{-1}, \bm{x}_k\}$
    }

    // Fast Inference

    \For {each triplet $\{\bm{z}_k, |{\rm det} \mathcal{J}_{f_\theta}(\bm{z}_k)|^{-1}, \bm{x}_k\}$} {
        $q_\phi(\bm{z}_k|\bm{c}) \gets \sum^{M}_{m=1} w_m \mathcal{N}(\bm{z}_k;\mu_m(\bm{c}),\sigma^2_m(\bm{c}))$

        $p(\bm{x}_k|\bm{c}) \gets q_\phi(\bm{z}_k|\bm{c}) |{\rm det} \mathcal{J}_{f_\theta}(\bm{z}_k)|^{-1}$

    }
\end{algorithm}

\section{Experimental Evaluation}
\label{sec:experimental_evaluation}
\setlength{\tabcolsep}{3pt}
\begin{table}[t!]
\caption{\textbf{Quantitative comparisons over the stochastic human motion prediction metrics on Human3.6M and AMASS datasets.} Lower is better for all metrics except APD. The reported inference time is when a method finishes generating 50 prediction samples from receiving the past motion.}
\footnotesize
\centering
\makebox[\textwidth]{%
\begin{tabular}{lrrrrr|ccccc|r}
\hline
 & \multicolumn{5}{c|}{Human3.6M \citep{ionescu2013h36m}} & \multicolumn{5}{c|}{AMASS \citep{mahmood2019amass}} & \multicolumn{1}{c}{\multirow{2}{*}{\begin{tabular}[c]{@{}c@{}}Inference\\ Time[ms]$\downarrow$\end{tabular}}} \\
 & \multicolumn{1}{c}{APD$\uparrow$} & \multicolumn{1}{c}{ADE$\downarrow$} & \multicolumn{1}{c}{FDE$\downarrow$} & \multicolumn{1}{c}{MMADE$\downarrow$} & \multicolumn{1}{c|}{MMFDE$\downarrow$} & APD$\uparrow$ & ADE$\downarrow$ & FDE$\downarrow$ & MMADE$\downarrow$ & MMFDE$\downarrow$ & \multicolumn{1}{c}{} \\ \hline
HP-GAN~\citep{barsoum2018hp} & 7.214 & 0.858 & 0.867 & 0.847 & 0.858 & - & - & - & - & - & - \\
DSF~\citep{yuan2019diverse} & 9.330 & 0.493 & 0.592 & 0.550 & 0.599 & - & - & - & - & - & - \\
DeLiGAN~\citep{gurumurthy2017deligan} & 6.509 & 0.483 & 0.534 & 0.520 & 0.545 & - & - & - & - & - & - \\
GMVAE~\citep{dilokthanakul2016deep} & 6.769 & 0.461 & 0.555 & 0.524 & 0.566 & - & - & - & - & - & - \\
TPK~\citep{walker2017pose} & 6.723 & 0.461 & 0.560 & 0.522 & 0.569 & \multicolumn{1}{r}{9.283} & \multicolumn{1}{r}{0.656} & \multicolumn{1}{r}{0.675} & \multicolumn{1}{r}{0.658} & \multicolumn{1}{r|}{0.674} & 30.3 \\
MT-VAE~\citep{yan2018mt} & 0.403 & 0.457 & 0.595 & 0.716 & 0.883 & - & - & - & - & - & - \\
BoM~\citep{bhattacharyya2018accurate} & 6.265 & 0.448 & 0.533 & 0.514 & 0.544 & - & - & - & - & - & - \\
DLow~\citep{yuan2020dlow} & 11.741 & 0.425 & 0.518 & 0.495 & 0.531 & \multicolumn{1}{r}{13.170} & \multicolumn{1}{r}{0.590} & \multicolumn{1}{r}{0.612} & \multicolumn{1}{r}{0.618} & \multicolumn{1}{r|}{0.617} & 30.8 \\
MultiObj~\citep{ma2022multi} & 14.240 & 0.414 & 0.516 & \multicolumn{1}{c}{-} & \multicolumn{1}{c|}{-} & - & - & - & - & - & - \\
GSPS~\citep{mao2021generating} & 14.757 & 0.389 & 0.496 & 0.476 & 0.525 & \multicolumn{1}{r}{12.465} & \multicolumn{1}{r}{0.563} & \multicolumn{1}{r}{0.613} & \multicolumn{1}{r}{0.609} & \multicolumn{1}{r|}{0.633} & 5.1 \\
Motron~\citep{salzmann2022motron} & 7.168 & 0.375 & 0.488 & 0.509 & 0.539 & - & - & - & - & - & - \\
DivSamp~\citep{dang2022diverse} & \textbf{15.310} & 0.370 & 0.485 & 0.475 & 0.516 & \multicolumn{1}{r}{\textbf{24.724}} & \multicolumn{1}{r}{0.564} & \multicolumn{1}{r}{0.647} & \multicolumn{1}{r}{0.623} & \multicolumn{1}{r|}{0.667} & 5.2 \\
BeLFusion~\citep{barquero2023belfusion} & 7.602 & 0.372 & 0.474 & 0.473 & 0.507 & \multicolumn{1}{r}{9.376} & \multicolumn{1}{r}{0.513} & \multicolumn{1}{r}{0.560} & \multicolumn{1}{r}{0.569} & \multicolumn{1}{r|}{0.585} & 449.3 \\
BeLFusion-D & 5.777 & 0.367 & 0.472 & 0.469 & 0.506 & \multicolumn{1}{r}{7.458} & \multicolumn{1}{r}{0.508} & \multicolumn{1}{r}{0.567} & \multicolumn{1}{r}{0.564} & \multicolumn{1}{r|}{0.591} & 39.3 \\
HumanMAC~\citep{chen2023humanmac} & 6.301 & 0.369 & 0.480 & 0.509 & 0.545 & \multicolumn{1}{r}{9.321} & \multicolumn{1}{r}{0.511} & \multicolumn{1}{r}{0.554} & \multicolumn{1}{r}{0.593} & \multicolumn{1}{r|}{0.591} & 1172.9 \\
CoMusion~\citep{sun2024comusion} & 7.632 & 0.350 & 0.458 & 0.494 & 0.506 & \multicolumn{1}{r}{10.848} & \multicolumn{1}{r}{\textbf{0.494}} & \multicolumn{1}{r}{\textbf{0.547}} & \multicolumn{1}{r}{\textbf{0.469}} & \multicolumn{1}{r|}{\textbf{0.466}} & 352.6 \\
SLD~\citep{xu2024learning} & 8.741 & 0.348 & \textbf{0.436} & \textbf{0.435} & \textbf{0.463} & - & - & - & - & - & 375.0 \\
\rev{SkeletonDiff.}\citep{curreli2025nonisotropic} & 7.632 & \textbf{0.344} & 0.450 & 0.487 & 0.512 & \multicolumn{1}{r}{9.456} & \multicolumn{1}{r}{0.480} & \multicolumn{1}{r}{0.545} & \multicolumn{1}{r}{0.561} & \multicolumn{1}{r|}{0.580} & 1013.5 \\ \hline
\methodname{} & 6.101 & 0.369 & 0.473 & 0.481 & 0.511 & \multicolumn{1}{r}{7.099} & \multicolumn{1}{r}{0.511} & \multicolumn{1}{r}{0.566} & \multicolumn{1}{r}{0.567} & \multicolumn{1}{r|}{0.586} & \textbf{1.3} \\
\multicolumn{1}{r}{w/o Cache} & 5.385 & 0.374 & 0.489 & 0.490 & 0.531 & \multicolumn{1}{r}{6.291} & \multicolumn{1}{r}{0.516} & \multicolumn{1}{r}{0.586} & \multicolumn{1}{r}{0.573} & \multicolumn{1}{r|}{0.608} & 415.9 \\ \hline
\end{tabular}}
\label{tab:quantitative}
\end{table}
\textbf{Datasets.}
We evaluate our \methodname{} on Human3.6M~\citep{ionescu2013h36m} and AMASS~\citep{mahmood2019amass}.
Human3.6M contains 3.6 million frames of human motion sequences.
Human motions of 11 subjects performing 15 actions are recorded at 50 Hz.
We follow the setting including the dataset split, the 16-joints pose skeleton definition, and lengths of past and future motions proposed by previous works~\citep{martinez2017simple,luvizon20182d,yang20183d,pavllo20193d}.
The training and test sets of Human3.6M are subjects [S1,S5,S6,S7,S8] and [S9,S11], respectively.
The past motion and future motions contain 25 frames (0.5 sec) and 100 frames (2.0 sec).
AMASS unifies 24 different human motion datasets including HumanEva-I~\citep{sigal2010humaneva} with the SMPL~\citep{SMPL} pose representation.
AMASS contains 9M frames at 60 Hz in total.
As a multi-dataset collection of AMASS, one can perform a cross-dataset evaluation.
We follow the evaluation protocol proposed by BeLFusion~\citep{barquero2023belfusion} for fair comparison, as predicting future 120 frames (2.0 sec) with 30 frames observation (0.5 sec) with downsampling to 60 Hz.

\textbf{Metrics.}
We use the evaluation metrics to measure diversity and accuracy.
50 sampled predictions are evaluated with the following metrics:
\textbf{Average Pairwise Distance (APD)}~\citep{aliakbarian2020stochastic} evaluates sample diversity.
It calculates the mean $l_2$ distance between all predicted motions.
\textbf{Average and Final Displacement Error (ADE, FDE)}~\citep{alahi2016social,lee2017desire,gupta2018social} evaluate accuracy.
They calculate the average and final-frame $l_2$ distances between the ground truth motion and closest prediction in the 50 set.
\textbf{Multimodal ADE and FDE (MMADE, MMFDE)}~\citep{yuan2019diverse} also evaluate accuracy in a similar way to ADE and FDE.
However, they are calculated over multimodal ground truths selected by grouping similar motions.

We also evaluate the accuracy of density estimation with \textbf{Multimodal Log Probability} per dimension.
It calculates the log probability of the multimodal ground truths to measure how accurately the estimated density covers possible future motions.
We evaluate the log probability on the motion space except for methods with latent space such as our \methodname{} and BeLFusion.
While higher is better on APD and multimodal log probability, lower is better on ADE, FDE, MMADE, and MMFDE.

\textbf{Implementation Details.}
Our method is based on a latent flow-based model as mentioned in Sec.~\ref{subsec:normalizing_flow}.
We utilize a Variational Autoencoder (VAE) \rev{for $\mathcal{E}, \mathcal{D}$} to obtain a latent representation.
Specifically, we employ the Behavioral Latent Space (BLS) \citep{barquero2023belfusion} as a VAE to achieve a compact latent representation.
BLS ensures smoothness of predicted motions and consistency between the end of the past motion and the start of the predicted motion.
Additionally, we compress this representation using linear factorization \citep{xu2024learning}.
The dimensionality of the VAE latent space is 128, which we further reduce to 8 dimensions through linear factorization.
We trained the unconditional flow-based model on this 8-dimensional space.
The unconditional flow-based model \( f_\theta \) is a continuous normalizing flow (CNF) model, with its vector field regressed by a U-Net architecture.
The conditional base density \( q_\phi \), as well as the VAE encoder and decoder, are implemented as one-layer Recurrent Neural Networks (RNNs).
We used a Gaussian mixture model with $M=50$ modes to model the conditional base density $q_\phi$.
We precomputed and collected triplets $t_k = \{\bm{z}_k, |{\rm det} \mathcal{J}_{f_\theta}(\bm{z}_k)|^{-1},  \bm{x}_k\}$ using all training samples of each dataset.
All experiments, including inference time measuring, were carried out using a single NVIDIA A100 GPU. We used a batch size 64 and the Adam optimizer with a learning rate of \( 5 \times 10^{-4} \).

\subsection{Quantitative Evaluation}


\textbf{Accuracy Over a Fixed Number of Predictions.}
We compare \methodname{} against state-of-the-art methods of stochastic human motion prediction.
While we propose using a precomputed set during inference, we also evaluate our method without precomputation.
In the absence of precomputation, we sample \(\bm{z}\) from the conditional base density \(q_\phi(\bm{z}|\bm{c})\) and obtain \(\bm{x}\) through the flow-based model inference, where \(\bm{x} = f_\theta(\bm{z})\).
The results are summarized in Table \ref{tab:quantitative}.
Since the primary applications of human motion prediction are in real-time scenarios, we also measure the inference time of each method to sample 50 predictions on a GPU.

CoMusion and SLD were successful in predicting motions that are closer to the ground truth than \methodname{}; however, their inference times of 167 
and 375 milliseconds are too long for the intended 2000 ms prediction horizon.
As a result, over 8\% of the first prediction sequence is rendered useless once the prediction is finalized.
Therefore, it is difficult to use these methods with slow inference in real-time applications.
Although our primary goal is to estimate the density, \methodname{} achieves comparable performances with a 1.3 millisecond inference time.
Our method achieves around 4$\times$ faster than the fastest VAE method, GSPS, and 30$\times$ faster than the fastest diffusion-based method, BeLFusion-D.
The inference of our method is fast enough (1.3ms for future 2000ms) and applicable for real-time applications.
This inference speed is because the inference of the unconditional flow-based model $f_\theta$ is precomputed.
We only need to evaluate the lightweight conditional base density $q_\phi$ at inference.
Although our conditional base density $q_\phi$ is just a Gaussian mixture with low expressive power, our method achieves high accuracy since the precomputed unconditional flow-based model $f_\theta$ gives $q_\phi$ much complexity with almost no overhead in inference.

\textbf{Density Estimation Accuracy.}
\begin{table}[t]
\caption{\textbf{Density Estimation Accuracy on Human3.6M.} Inference time of each method is reported as \{total time (time without KDE inference)\}. Since our method doesn't require KDE for density estimation, the number of samples for KDE is left blank for \methodname{}.}
    \centering
    \begin{tabular}{lrrrr}
\toprule
\multicolumn{1}{c}{Method} & \multicolumn{1}{c}{\begin{tabular}[c]{@{}c@{}}\#sample\\ for KDE\end{tabular}} & \multicolumn{1}{c}{\begin{tabular}[c]{@{}c@{}}MM log prob.\\ per dim $\uparrow$\end{tabular}} & \multicolumn{2}{c}{\begin{tabular}[c]{@{}c@{}}Inference\\ Time[ms] $\downarrow$\end{tabular}} \\ 
\midrule
BeLFusion~\citep{barquero2023belfusion} & 50 & -2.383 & 2305.3 & (440.3) \\
 & 1000 & -1.633 & 2422.4 & (449.3) \\ 
CoMusion~\citep{sun2024comusion} & 50 & -15.575 & 2500.5 & (167.0) \\
 & 1000 & -12.746 & 5071.5 & (2741.3) \\
SLD~\citep{xu2024learning} & 50 & 0.080 & 2559.1 & (375.0) \\ 
\midrule
\methodname & - & \textbf{1.304} & \textbf{0.5} & (0.5) \\ 
\bottomrule
\end{tabular}
\label{tab:probability}
\end{table}
\begin{figure}
    \centering
    \includegraphics[width=\textwidth]{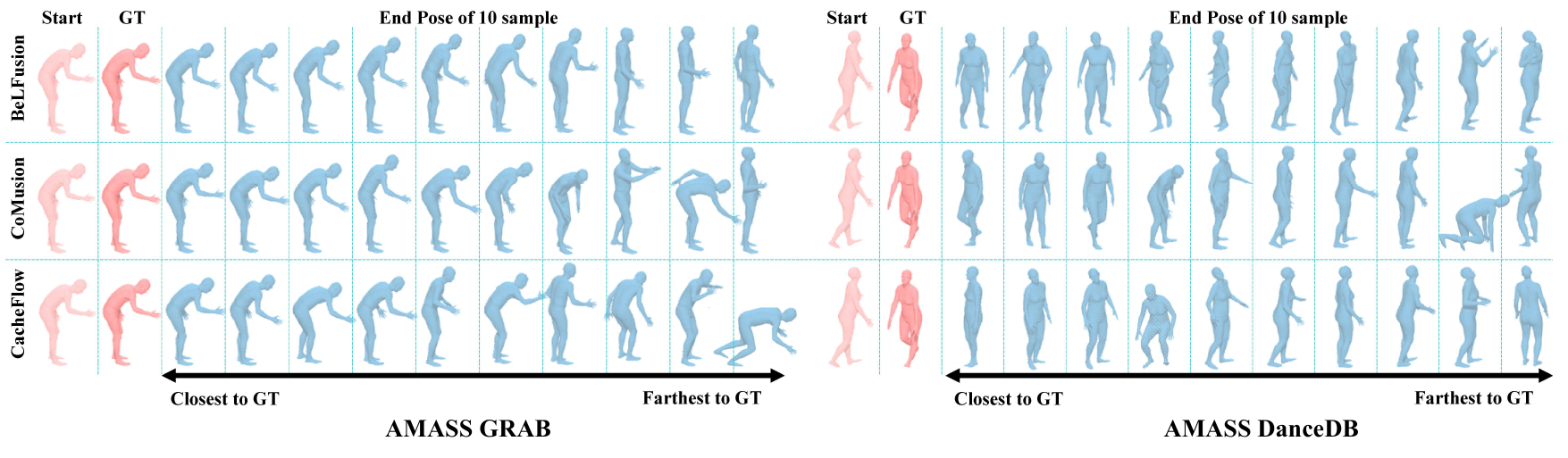}
    \caption{\textbf{Qualitative Comparison on AMASS dataset.}}
    \label{fig:qualitative_comparison}
\end{figure}
The density estimation accuracy of each method is compared between \methodname{} and the state-of-the-art methods.
The three state-of-the-art methods BeLFusion~\citep{barquero2023belfusion}, CoMusion~\citep{sun2024comusion}, and SLD~\citep{xu2024learning} are selected.
CoMusion and SLD were selected since they outperform our method in benchmarks of stochastic human motion prediction.
We also include BeLFusion to compare \methodname{} with the method with latent space.
We applied KDE to these previous methods since they only sample a set of predictions and cannot estimate density.
While we evaluated 50 and 1000 samples for KDE on BeLFusion and CoMusion, SLD only allows 50 samples due to the fixed number of anchors corresponding to predictions.
We measured the inference time of each method to estimate the density of ten thousand future motions from the past motion input.

The quantitative comparisons over the multimodal ground truth log probability are shown in \cref{tab:probability}.
All previous methods suffer from slow inference of their own and KDE on high-dimensional motion data.
Their inference time exceeded the prediction horizon of 2000ms in the future.
Therefore, they cannot estimate density in real-time.
In contrast, our method achieves better estimation accuracy in less than one millisecond.
This indicates that \methodname{} has strong discriminative ability to list up possible future motions required for safety assurance.
Our method is even faster only on the density estimation (0.5ms) than the inference time reported in \cref{tab:quantitative} (1.3ms).
This is because we don't need any extra sampling operation in the density estimation.

\subsection{Qualitative Comparison of Predicted Motions}

To visually evaluate \methodname{}, we conducted a qualitative comparison of methods on the AMASS dataset, as shown in \cref{fig:qualitative_comparison}.
We visualized the end poses of 10 samples from each method alongside the end poses of past motions and the ground truth future motions.
The sitting or lying poses were translated to the ground plane, as the global translation is not modeled in human motion prediction.
The 10 pose samples are arranged from the closest to the farthest from the ground truth pose based on joint rotations.

Our observations indicate that \methodname{} predicts realistic poses.
The closest poses to the ground truths also demonstrate that the accuracy of \methodname{} is comparable to CoMusion, as reflected in the ADE and FDE metrics listed in \cref{tab:quantitative}.
Notably, our method is computationally efficient, operating 100 times faster than the fastest CoMusion.
In summary, \methodname{} effectively delivers realistic and accurate predictions.

\subsection{\rev{Further Analysis}}

\subsubsection{Generalization to Out-of-Distribution Data}

\begin{table}[t]
\caption{\textbf{Zero-shot evaluation on 3DPW~\citep{3DPW}.} The models are trained on AMASS~\citep{mahmood2019amass}. For \methodname{}, the cache are computed on AMASS training set also.}
\centering
\begin{tabular}{lrrrrr}
\hline
 & \multicolumn{5}{c}{Zero-shot 3DPW~\citep{3DPW}} \\
 & \multicolumn{1}{c}{APD$\uparrow$} & \multicolumn{1}{c}{ADE$\downarrow$} & \multicolumn{1}{c}{FDE$\downarrow$} & \multicolumn{1}{c}{MMADE$\downarrow$} & \multicolumn{1}{c}{MMFDE$\downarrow$} \\ \hline
TPK~\citep{walker2017pose} & 9.582 & 0.648 & 0.701 & 0.665 & 0.702 \\
DLow~\citep{yuan2020dlow} & 13.772 & 0.581 & 0.649 & 0.602 & 0.651 \\
GSPS~\citep{mao2021generating} & 11.809 & 0.552 & 0.650 & 0.578 & 0.653 \\
DivSamp~\citep{dang2022diverse} & \textbf{24.153} & 0.554 & 0.678 & 0.593 & 0.686 \\
BeLFusion~\citep{barquero2023belfusion} & 7.740 & 0.493 & 0.590 & 0.531 & 0.599 \\
CoMusion~\citep{sun2024comusion} & 11.404 & 0.477 & 0.570 & 0.540 & 0.587 \\
SkeletonDiff.\citep{curreli2025nonisotropic} & 9.814 & \textbf{0.472} & 0.575 & 0.535 & 0.594 \\ \hline
\methodname{} & 7.963 & 0.499 & 0.591 & 0.539 & 0.601 \\ \hline
\end{tabular}
\label{exp:zero-shot}
\end{table}

To evaluate the generalization capability of \methodname{} on unseen domains, we conduct a zero-shot evaluation on the 3DPW dataset~\citep{3DPW}. In this setting, all models are trained strictly on the AMASS dataset~\citep{mahmood2019amass} and tested directly on 3DPW without any fine-tuning. This poses a significant challenge due to the domain shift between the indoor optical motion capture data of AMASS and the in-the-wild outdoor sequences of 3DPW.

The results are summarized in Tab.~\ref{exp:zero-shot}. Despite the challenging zero-shot setting, \methodname{} demonstrates robust generalization performance. Specifically, our method achieves an ADE of 0.499 and an FDE of 0.591, surpassing earlier baselines such as TPK, DLow, and GSPS. Furthermore, our performance is highly competitive with recent diffusion-based approaches, such as BeLFusion~\citep{barquero2023belfusion} (0.493 ADE) and CoMusion~\citep{sun2024comusion} (0.477 ADE), showing only marginal differences in error metrics.

A crucial aspect of this experiment is the configuration of the cache. As noted in the table, the cache used by \methodname{} is constructed solely from the AMASS training set. The competitive performance on 3DPW indicates that \methodname{} does not simply memorize training samples. Instead, it successfully leverages the retrieved motion priors from the source domain (AMASS) to reconstruct valid motion trajectories in the target domain (3DPW). This suggests that the latent flow capabilities of \methodname{} effectively bridge the domain gap by generalizing the transitions found in the cached training data to unseen, in-the-wild scenarios.

\methodname{} generalizes beyond training data in two key ways. First, although cached latents $z_k$ are drawn from the training set, generated futures $\hat{X}$ differ significantly because they are conditioned on novel past contexts $c$. This allows the model to generate motions unseen during training. Second, the large scale of the Human3.6M (371,188 patterns) and AMASS (723,263 patterns) datasets already provides considerable diversity, and \methodname{} leverages this diversity to further expand to novel futures. Even with fixed cached codes, novel inputs yield generalized future motions that go beyond memorization of the training set.  

\subsubsection{Extensive Computational and Memory Cost Analysis}
\label{subsubsec:cost}
We conducted a detailed evaluation of both computational overhead and memory requirements for \textbf{\methodname{}} across motion prediction benchmarks, using an Intel(R) Xeon(R) Gold 6154 CPU and a single NVIDIA TITAN RTX GPU. The results demonstrate that our caching mechanism achieves efficient memory usage while maintaining practical inference times on both CPU and low-performance GPU platforms.  

\[
\text{Memory Cost (MB)} = \text{Bytes per motion} \times \#\text{motion} \,/\, 1024^2
\]  

The measured costs are summarized in Table~\ref{tab:cost}.  

\begin{table}[t]
\caption{Computational and memory cost analysis on Human3.6M and AMASS datasets.}
\centering
\begin{tabular}{l|llll|llll}
Cost & Memory &  &  &  & Computation &  &  &  \\ \hline
 & \multirow{2}{*}{\begin{tabular}[c]{@{}l@{}}Bytes\\ per Mot.\end{tabular}} & \multirow{2}{*}{\# Mot.} & \multirow{2}{*}{\begin{tabular}[c]{@{}l@{}}Cache Size\\ (MB)\end{tabular}} & \multirow{2}{*}{\begin{tabular}[c]{@{}l@{}}GPU mem.\\ (MB)\end{tabular}} & GPU &  & CPU &  \\
 &  &  &  &  & Cache & Inf. & Cache & Inf \\
Human3.6M & \multicolumn{1}{r}{68} & \multicolumn{1}{r}{371188} & \multicolumn{1}{r}{24.07} & \multicolumn{1}{r|}{811} & \multicolumn{1}{r}{26min} & \multicolumn{1}{r}{2.7ms} & \multicolumn{1}{r}{1h19min} & \multicolumn{1}{r}{3.6ms} \\
AMASS & \multicolumn{1}{r}{68} & \multicolumn{1}{r}{723263} & \multicolumn{1}{r}{46.90} & \multicolumn{1}{r|}{875} & \multicolumn{1}{r}{1h10min} & \multicolumn{1}{r}{3.4ms} & \multicolumn{1}{r}{3h40min} & \multicolumn{1}{r}{6.9ms}
\end{tabular}
\label{tab:cost}
\end{table}

This analysis highlights three important aspects. First, the cache footprint is very small, less than 50 MB for both datasets. Second, the preprocessing cost is modest, requiring only one to four hours even for AMASS on CPU. Third, the inference time is highly practical, amounting to only a few milliseconds per sample on both CPU and GPU, which makes \methodname{} suitable for real-time applications.  

\subsubsection{Ablation Study}
\begin{table}[t]
 \caption{\textbf{Ablation Study on Human3.6M.} (4) and (5) do not affect the ground truth log probability, these are left blank.}
    \footnotesize
    \centering
    \makebox[\textwidth]{%
    \begin{tabular}{lccccc|cccc}
\toprule
\multicolumn{1}{c}{} & \begin{tabular}[c]{@{}c@{}}Linear\\ Factorization\end{tabular} & \begin{tabular}[c]{@{}c@{}}Unconditional\\ Flow-based Model\end{tabular} & \begin{tabular}[c]{@{}c@{}}Joint\\ Learning\end{tabular} & Precomp. Set & Sampling & ADE$\downarrow$ & FDE$\downarrow$ & \begin{tabular}[c]{@{}c@{}}MM log prob.\\ per dims$\uparrow$\end{tabular} & \begin{tabular}[c]{@{}c@{}}Inference\\ time[ms]$\downarrow$\end{tabular} \\ 
\midrule
(1) &  & \checkmark & \checkmark & Train Set & NN sample & 0.502 & 0.664 & 0.458 & 4.8 \\
(2) & \checkmark & \multicolumn{1}{l}{} & \checkmark & Train Set & NN sample & 0.616 & 0.889 & 0.901 & 0.4 \\
(3) & \checkmark & \checkmark &  & Train Set & NN sample & 0.370 & 0.475 & 1.283 & 1.3 \\ \cmidrule{1-6}
(4) & \checkmark & \checkmark & \checkmark & Base Density & NN sample & 0.376 & 0.492 & - & 1.3 \\ \cmidrule{1-6}
\multirow{2}{*}{(5)} & \checkmark & \checkmark & \checkmark & Train Set & Random sample & 0.455 & 0.605 & - & \textbf{1.2} \\
 & \checkmark & \checkmark & \checkmark & Train Set & Most likely & 0.384 & 0.506 & - & 1.4 \\ \cmidrule{1-6}
 & \checkmark & \checkmark & \checkmark & Train Set & NN sample & \textbf{0.369} & \textbf{0.473} & \textbf{1.304} & 1.3 \\
 \bottomrule
 \end{tabular}}
 \label{tab:ablation}
 \vspace{-4mm}
\end{table}
We conducted an ablation study to investigate how each component affects the performance of our \methodname{}.
We ablate five components: (1) dimensionality reduction via linear factorization on VAE, (2) the unconditional flow-based model $f_\theta$, (3) joint learning of the conditional base density $q_\phi$ and unconditional flow-based model $f_\theta$, (4) dataset for precomputation, (5) the sampling method for metrics over a fixed number of predictions.
Ablation results on the Human3.6M dataset are summarized in \cref{tab:ablation}.

\noindent \textbf{Linear Factorization.} We first ablate the linear factorization compressing 256-dim VAE latent to be an 8-dim factor space.
Our method is considerably enhanced on the compact space by avoiding the curse of dimensionality.

\noindent \textbf{The Unconditional Flow-based Model.} We ablate this flow-based model $f_\theta$ to confirm it improves the conditional base density $q_\phi$ by adding complexity. 
As shown in \cref{tab:ablation}, we observe a notable performance drop without the flow-based model.
Therefore, our unconditional flow-based model $f_\theta$ complements conditional base density $q_\phi$ to estimate complex density distribution over human motions.

\noindent \textbf{Joint Learning.} We ablate the joint learning of the unconditional flow-based model $f_\theta$ and the conditional base density $q_\phi$.
The joint learning certainly improves both prediction errors and density estimation accuracy.
The unconditional flow-based model $f_\theta$ can learn a more clustered $\bm{z}$ mapped from the motion feature $\bm{x}$.
Thus, a conditional base density $q_\phi$ can easily model the $\bm{z}$ distribution.

\noindent \textbf{Cache Precomputation.}
We propose the precomputation over the training split.
Specifically, we apply inverse transform $\bm{z} = f_\theta(\bm{x})$ to ground truth future motions in the training split.
However, we may precompute infinite precomputation samples.
For example, we can sample $\bm{z} \sim q_\phi(\bm{z}|\bm{c})$ and obtain $\bm{x}$ by forward transform $\bm{x} = f_\theta(\bm{z})$.
As shown in the ablation, precomputation on the training split outperforms one on the base density since we can regularize the prediction to be legitimate human motions using the training split.

\noindent \textbf{Sampling Method.}
We propose the nearest neighbor sampling from the precomputation set as described in \cref{subsec:efficient_density}.
Lastly, we ablate this sampling to evaluate its performance gain.
We experimented with two sampling method alternatives: random sampling and most likely sampling.
Precomputed motion features $\bm{x}_{k^*}$ are uniformly selected as predictions with random sampling.
Most likely sampling selects motion features $\bm{x}_{k^*}$ with the highest probabilities $k^* = {\rm argmax}_k p(\bm{x}_k|\bm{c})$.
We found that the large and little performance drops with random and most likely sampling respectively.
This random sampling is worse due to the independence from the past motions $\bm{c}$.
The most likely method underperforms due to less diverse samples. It cannot select a motion feature set with diversity because all selected features are often located in one peak of the estimated density.
Since ADE and FDE are best-of-many metrics, this less diversity leads to worse performance.
In contrast, our sampling method is superior to others.
Our sampling incorporates past motions and achieves good diversity by simulating sampling from the estimated density $p(\bm{x}|\bm{c})$.

\section{Concluding Remarks}
\label{sec:conclusion}

We presented a new flow-based stochastic human motion prediction method named \methodname{}.
Our method achieves a fast and accurate estimation of the probability density distribution of future motions.
Our unconditional formulation allows precomputation and caching of the flow-based model, thus omitting a large portion of computational cost at inference.
The unconditional flow-based model enhanced the expressivity of the lightweight conditional Gaussian mixture with almost no overhead.
Experimental results demonstrated  \methodname{} achieved comparable prediction accuracy with 1.3 milliseconds inference, much faster than the previous method.
Furthermore, \methodname{} estimated a more accurate density than previous methods in less than 1 millisecond.

Our method has one limitation.
Prediction and density estimation are performed within precomputed triplets.
We cannot estimate the density or predict unseen future motions during precomputation.
Our future work is searching for a better precomputation strategy for prediction and estimation with more coverage based on the limited dataset.
Furthermore, our method is not limited to prediction tasks but applies to any regression task requiring density estimation.
We will investigate the applicability of our \methodname{} on other domains.

\clearpage

\bibliography{venues_string, tmlr}

@article{survey_autonomous_driving,
  title={A survey of motion planning and control techniques for self-driving urban vehicles},
  author={Paden, Brian and {\v{C}}{\'a}p, Michal and Yong, Sze Zheng and Yershov, Dmitry and Frazzoli, Emilio},
  journal={IEEE Transactions on intelligent vehicles},
  volume={1},
  number={1},
  pages={33--55},
  year={2016}
}

@inproceedings{butepage2018anticipating,
  title={Anticipating many futures: Online human motion prediction and generation for human-robot interaction},
  author={B{\"u}tepage, Judith and Kjellstr{\"o}m, Hedvig and Kragic, Danica},
  booktitle= ICRA,
  year={2018},
}

@article{koppula2015anticipating,
  title={Anticipating human activities using object affordances for reactive robotic response},
  author={Koppula, Hema S and Saxena, Ashutosh},
  journal= PAMI,
  volume={38},
  number={1},
  pages={14--29},
  year={2015},
}

@inproceedings{lasota2017multiple,
  title={A multiple-predictor approach to human motion prediction},
  author={Lasota, Przemyslaw A and Shah, Julie A},
  booktitle= ICRA,
  pages={2300--2307},
  year={2017}
}

@inproceedings{mainprice2013human,
  title={Human-robot collaborative manipulation planning using early prediction of human motion},
  author={Mainprice, Jim and Berenson, Dmitry},
  booktitle={2013 IEEE/RSJ International Conference on Intelligent Robots and Systems},
  pages={299--306},
  year={2013},
  organization={IEEE}
}

@inproceedings{sanderud2015likelihood,
  title={A likelihood analysis for a risk analysis for safe human robot collaboration},
  author={Sanderud, Audun R{\o}nning and Niitsuma, Mihoko and Thomessen, Trygve},
  booktitle={2015 IEEE 20th Conference on Emerging Technologies \& Factory Automation (ETFA)},
  pages={1--6},
  year={2015},
  organization={IEEE}
}

@inproceedings{tisnikar2024probabilistic,
  title={Probabilistic Inference of Human Capabilities from Passive Observations},
  author={Tisnikar, Peter and Canal, Gerard and Leonetti, Matteo},
  booktitle={2024 IEEE/RSJ International Conference on Intelligent Robots and Systems (IROS)},
  pages={8779--8785},
  year={2024},
  organization={IEEE}
}

@article{vae,
  title={Auto-encoding variational bayes},
  author={Kingma, Diederik P and Welling, Max},
  journal= ICLR,
  year={2014}
}

@article{gan,
  title={Generative adversarial networks},
  author={Goodfellow, Ian and Pouget-Abadie, Jean and Mirza, Mehdi and Xu, Bing and Warde-Farley, David and Ozair, Sherjil and Courville, Aaron and Bengio, Yoshua},
  journal={Communications of the ACM},
  volume={63},
  number={11},
  pages={139--144},
  year={2020},
}

@article{ddpm,
  title={Denoising diffusion probabilistic models},
  author={Ho, Jonathan and Jain, Ajay and Abbeel, Pieter},
  journal= NeurIPS,
  volume={33},
  pages={6840--6851},
  year={2020}
}

@article{rosenblatt1956remarks,
  title={Remarks on some nonparametric estimates of a density function},
  author={Rosenblatt, Murray},
  journal={The annals of mathematical statistics},
  pages={832--837},
  year={1956},
}

@article{parzen1962estimation,
  title={On estimation of a probability density function and mode},
  author={Parzen, Emanuel},
  journal={The annals of mathematical statistics},
  volume={33},
  number={3},
  pages={1065--1076},
  year={1962},
}

@book{silverman2018density,
  author       = {Bernard W. Silverman},
  title        = {Density Estimation for Statistics and Data Analysis},
  publisher    = {Springer},
  year         = {1986},
}

@article{DBLP:journals/corr/abs-1802-03426,
  author       = {Leland McInnes and
                  John Healy},
  title        = {{UMAP:} Uniform Manifold Approximation and Projection for Dimension
                  Reduction},
  journal      = {CoRR},
  volume       = {abs/1802.03426},
  year         = {2018},
  url          = {http://arxiv.org/abs/1802.03426},
  eprinttype    = {arXiv},
  eprint       = {1802.03426},
  bibsource    = {dblp computer science bibliography, https://dblp.org}
}

@inproceedings{rezende2015variational,
  title={Variational inference with normalizing flows},
  author={Rezende, Danilo and Mohamed, Shakir},
  booktitle= ICML,
  pages={1530--1538},
  year={2015}
}

@article{tabak2013family,
  title={A family of nonparametric density estimation algorithms},
  author={Tabak, Esteban G and Turner, Cristina V},
  journal={Communications on Pure and Applied Mathematics},
  volume={66},
  number={2},
  pages={145--164},
  year={2013},
}

@article{tabak2010density,
  title={Density estimation by dual ascent of the log-likelihood},
  author={Tabak, Esteban G and Vanden-Eijnden, Eric},
  journal={Communications in Mathematical Sciences},
  volume={8},
  number={1},
  pages={217--233},
  year={2010},
}

@article{chen2018neural,
  title={Neural ordinary differential equations},
  author={Chen, Ricky TQ and Rubanova, Yulia and Bettencourt, Jesse and Duvenaud, David K},
  journal= NeurIPS,
  volume={31},
  year={2018}
}

@inproceedings{DBLP:conf/iclr/LipmanCBNL23,
  author       = {Yaron Lipman and
                  Ricky T. Q. Chen and
                  Heli Ben{-}Hamu and
                  Maximilian Nickel and
                  Matthew Le},
  title        = {Flow Matching for Generative Modeling},
  booktitle    =  ICLR,
  year         = {2023},
}

@inproceedings{DBLP:conf/iclr/LiuG023,
  author       = {Xingchao Liu and
                  Chengyue Gong and
                  Qiang Liu},
  title        = {Flow Straight and Fast: Learning to Generate and Transfer Data with
                  Rectified Flow},
  booktitle    =  ICLR,
  year         = {2023},
}

@article{conditional_normalizing_flow,
  author       = {Christina Winkler and
                  Daniel E. Worrall and
                  Emiel Hoogeboom and
                  Max Welling},
  title        = {Learning Likelihoods with Conditional Normalizing Flows},
  journal      = {CoRR},
  volume       = {abs/1912.00042},
  year         = {2019},
  url          = {http://arxiv.org/abs/1912.00042},
  eprinttype    = {arXiv},
  eprint       = {1912.00042},
  bibsource    = {dblp computer science bibliography, https://dblp.org}
}

@inproceedings{DBLP:conf/iclr/GrathwohlCBSD19,
  author       = {Will Grathwohl and
                  Ricky T. Q. Chen and
                  Jesse Bettencourt and
                  Ilya Sutskever and
                  David Duvenaud},
  title        = {{FFJORD:} Free-Form Continuous Dynamics for Scalable Reversible Generative
                  Models},
  booktitle    =  ICLR,
  year         = {2019},
  url          = {https://openreview.net/forum?id=rJxgknCcK7},
  bibsource    = {dblp computer science bibliography, https://dblp.org}
}

@inproceedings{maeda2023fast,
  title={Fast inference and update of probabilistic density estimation on trajectory prediction},
  author={Maeda, Takahiro and Ukita, Norimichi},
  booktitle= ICCV,
  pages={9795--9805},
  year={2023}
}

@article{ionescu2013h36m,
  title={Human3. 6m: Large scale datasets and predictive methods for 3d human sensing in natural environments},
  author={Ionescu, Catalin and Papava, Dragos and Olaru, Vlad and Sminchisescu, Cristian},
  journal= PAMI,
  volume={36},
  number={7},
  pages={1325--1339},
  year={2013}
}

@inproceedings{martinez2017simple,
  title={A simple yet effective baseline for 3d human pose estimation},
  author={Martinez, Julieta and Hossain, Rayat and Romero, Javier and Little, James J},
  booktitle= ICCV,
  pages={2640--2649},
  year={2017}
}

@inproceedings{luvizon20182d,
  title={2d/3d pose estimation and action recognition using multitask deep learning},
  author={Luvizon, Diogo C and Picard, David and Tabia, Hedi},
  booktitle= CVPR,
  pages={5137--5146},
  year={2018}
}

@inproceedings{yang20183d,
  title={3d human pose estimation in the wild by adversarial learning},
  author={Yang, Wei and Ouyang, Wanli and Wang, Xiaolong and Ren, Jimmy and Li, Hongsheng and Wang, Xiaogang},
  booktitle= CVPR,
  pages={5255--5264},
  year={2018}
}

@inproceedings{pavllo20193d,
  title={3d human pose estimation in video with temporal convolutions and semi-supervised training},
  author={Pavllo, Dario and Feichtenhofer, Christoph and Grangier, David and Auli, Michael},
  booktitle= CVPR,
  pages={7753--7762},
  year={2019}
}

@inproceedings{mahmood2019amass,
  title={AMASS: Archive of motion capture as surface shapes},
  author={Mahmood, Naureen and Ghorbani, Nima and Troje, Nikolaus F and Pons-Moll, Gerard and Black, Michael J},
  booktitle= ICCV,
  pages={5442--5451},
  year={2019}
}

@inproceedings{3DPW,
  title={Recovering accurate 3d human pose in the wild using imus and a moving camera},
  author={Von Marcard, Timo and Henschel, Roberto and Black, Michael J and Rosenhahn, Bodo and Pons-Moll, Gerard},
  booktitle= ECCV,
  pages={601--617},
  year={2018}
}

@article{SMPL,
  author       = {Matthew Loper and
                  Naureen Mahmood and
                  Javier Romero and
                  Gerard Pons{-}Moll and
                  Michael J. Black},
  title        = {{SMPL:} a skinned multi-person linear model},
  journal      = {{ACM} Trans. Graph.},
  volume       = {34},
  number       = {6},
  pages        = {248:1--248:16},
  year         = {2015}
}

@article{sigal2010humaneva,
  title={Humaneva: Synchronized video and motion capture dataset and baseline algorithm for evaluation of articulated human motion},
  author={Sigal, Leonid and Balan, Alexandru O and Black, Michael J},
  journal= IJCV,
  volume={87},
  number={1},
  pages={4--27},
  year={2010}
}

@inproceedings{aliakbarian2020stochastic,
  title={A stochastic conditioning scheme for diverse human motion prediction},
  author={Aliakbarian, Sadegh and Saleh, Fatemeh Sadat and Salzmann, Mathieu and Petersson, Lars and Gould, Stephen},
  booktitle= CVPR,
  pages={5223--5232},
  year={2020}
}

@inproceedings{alahi2016social,
  title={Social lstm: Human trajectory prediction in crowded spaces},
  author={Alahi, Alexandre and Goel, Kratarth and Ramanathan, Vignesh and Robicquet, Alexandre and Fei-Fei, Li and Savarese, Silvio},
  booktitle= CVPR,
  pages={961--971},
  year={2016}
}

@inproceedings{lee2017desire,
  title={Desire: Distant future prediction in dynamic scenes with interacting agents},
  author={Lee, Namhoon and Choi, Wongun and Vernaza, Paul and Choy, Christopher B and Torr, Philip HS and Chandraker, Manmohan},
  booktitle= CVPR,
  pages={336--345},
  year={2017}
}

@inproceedings{gupta2018social,
  title={Social gan: Socially acceptable trajectories with generative adversarial networks},
  author={Gupta, Agrim and Johnson, Justin and Fei-Fei, Li and Savarese, Silvio and Alahi, Alexandre},
  booktitle= CVPR,
  pages={2255--2264},
  year={2018}
}

@inproceedings{anchor,
  title={Diverse human motion prediction guided by multi-level spatial-temporal anchors},
  author={Xu, Sirui and Wang, Yu-Xiong and Gui, Liang-Yan},
  booktitle= ECCV,
  year={2022}
}

@inproceedings{aksan2021spatio,
  author       = {Emre Aksan and
                  Manuel Kaufmann and
                  Peng Cao and
                  Otmar Hilliges},
  title        = {A Spatio-temporal Transformer for 3D Human Motion Prediction},
  booktitle    = {International Conference on 3D Vision, 3DV 2021},
  pages        = {565--574},
  year         = {2021},
}

@inproceedings{bouazizi2022motionmixer,
  author       = {Arij Bouazizi and
                  Adrian Holzbock and
                  Ulrich Kressel and
                  Klaus Dietmayer and
                  Vasileios Belagiannis},
  title        = {MotionMixer: MLP-based 3D Human Body Pose Forecasting},
  booktitle    = {Proceedings of the Thirty-First International Joint Conference on
                  Artificial Intelligence, {IJCAI} 2022},
  pages        = {791--798},
  year         = {2022},
}

@inproceedings{fragkiadaki2015recurrent,
  title={Recurrent network models for human dynamics},
  author={Fragkiadaki, Katerina and Levine, Sergey and Felsen, Panna and Malik, Jitendra},
  booktitle= ICCV,
  pages={4346--4354},
  year={2015}
}

@inproceedings{butepage2017deep,
  title={Deep representation learning for human motion prediction and classification},
  author={Butepage, Judith and Black, Michael J and Kragic, Danica and Kjellstrom, Hedvig},
  booktitle= CVPR,
  pages={6158--6166},
  year={2017}
}

@inproceedings{guo2023back,
  title={Back to mlp: A simple baseline for human motion prediction},
  author={Guo, Wen and Du, Yuming and Shen, Xi and Lepetit, Vincent and Alameda-Pineda, Xavier and Moreno-Noguer, Francesc},
  booktitle= WACV,
  pages={4809--4819},
  year={2023}
}

@inproceedings{jain2016structural,
  title={Structural-rnn: Deep learning on spatio-temporal graphs},
  author={Jain, Ashesh and Zamir, Amir R and Savarese, Silvio and Saxena, Ashutosh},
  booktitle= CVPR,
  pages={5308--5317},
  year={2016}
}

@inproceedings{li2018convolutional,
  title={Convolutional sequence to sequence model for human dynamics},
  author={Li, Chen and Zhang, Zhen and Lee, Wee Sun and Lee, Gim Hee},
  booktitle= CVPR,
  pages={5226--5234},
  year={2018}
}

@inproceedings{martinez2017human,
  title={On human motion prediction using recurrent neural networks},
  author={Martinez, Julieta and Black, Michael J and Romero, Javier},
  booktitle= CVPR,
  pages={2891--2900},
  year={2017}
}

@inproceedings{gui2018adversarial,
  title={Adversarial geometry-aware human motion prediction},
  author={Gui, Liang-Yan and Wang, Yu-Xiong and Liang, Xiaodan and Moura, Jos{\'e} MF},
  booktitle= ECCV,
  pages={786--803},
  year={2018}
}

@article{pavllo2018quaternet,
  title={Quaternet: A quaternion-based recurrent model for human motion},
  author={Pavllo, Dario and Grangier, David and Auli, Michael},
  journal={BMVC},
  year={2018}
}

@inproceedings{liu2019towards,
  title={Towards natural and accurate future motion prediction of humans and animals},
  author={Liu, Zhenguang and Wu, Shuang and Jin, Shuyuan and Liu, Qi and Lu, Shijian and Zimmermann, Roger and Cheng, Li},
  booktitle= CVPR,
  pages={10004--10012},
  year={2019}
}

@inproceedings{medjaouri2022hr,
  title={Hr-stan: High-resolution spatio-temporal attention network for 3d human motion prediction},
  author={Medjaouri, Omar and Desai, Kevin},
  booktitle= CVPR,
  pages={2540--2549},
  year={2022}
}

@inproceedings{cai2020learning,
  title={Learning progressive joint propagation for human motion prediction},
  author={Cai, Yujun and Huang, Lin and Wang, Yiwei and Cham, Tat-Jen and Cai, Jianfei and Yuan, Junsong and Liu, Jun and Yang, Xu and Zhu, Yiheng and Shen, Xiaohui and others},
  booktitle= ECCV,
  pages={226--242},
  year={2020},
}

@inproceedings{martinez2021pose,
  title={Pose transformers (potr): Human motion prediction with non-autoregressive transformers},
  author={Mart{\'\i}nez-Gonz{\'a}lez, Angel and Villamizar, Michael and Odobez, Jean-Marc},
  booktitle= ICCV,
  pages={2276--2284},
  year={2021}
}

@inproceedings{li2020dynamic,
  title={Dynamic multiscale graph neural networks for 3d skeleton based human motion prediction},
  author={Li, Maosen and Chen, Siheng and Zhao, Yangheng and Zhang, Ya and Wang, Yanfeng and Tian, Qi},
  booktitle= CVPR,
  pages={214--223},
  year={2020}
}

@inproceedings{mao2019learning,
  title={Learning trajectory dependencies for human motion prediction},
  author={Mao, Wei and Liu, Miaomiao and Salzmann, Mathieu and Li, Hongdong},
  booktitle= ICCV,
  pages={9489--9497},
  year={2019}
}

@article{kobyzev2020normalizing,
  title={Normalizing flows: An introduction and review of current methods},
  author={Kobyzev, Ivan and Prince, Simon JD and Brubaker, Marcus A},
  journal= PAMI,
  volume={43},
  number={11},
  pages={3964--3979},
  year={2020},
  publisher={IEEE}
}

@inproceedings{dang2021msr,
  title={Msr-gcn: Multi-scale residual graph convolution networks for human motion prediction},
  author={Dang, Lingwei and Nie, Yongwei and Long, Chengjiang and Zhang, Qing and Li, Guiqing},
  booktitle= ICCV,
  pages={11467--11476},
  year={2021}
}

@inproceedings{li2021skeleton,
  title={Skeleton graph scattering networks for 3d skeleton-based human motion prediction},
  author={Li, Maosen and Chen, Siheng and Liu, Zihui and Zhang, Zijing and Xie, Lingxi and Tian, Qi and Zhang, Ya},
  booktitle= ICCV,
  pages={854--864},
  year={2021}
}

@inproceedings{barquero2023belfusion,
  title={Belfusion: Latent diffusion for behavior-driven human motion prediction},
  author={Barquero, German and Escalera, Sergio and Palmero, Cristina},
  booktitle= ICCV,
  pages={2317--2327},
  year={2023}
}

@inproceedings{xu2024learning,
  title={Learning semantic latent directions for accurate and controllable human motion prediction},
  author={Xu, Guowei and Tao, Jiale and Li, Wen and Duan, Lixin},
  booktitle= ECCV,
  pages={56--73},
  year={2024}
}

@inproceedings{sun2024comusion,
  title={CoMusion: Towards Consistent Stochastic Human Motion Prediction via Motion Diffusion},
  author={Sun, Jiarui and Chowdhary, Girish},
  booktitle= ECCV,
  pages={18--36},
  year={2024}
}

@inproceedings{barsoum2018hp,
  title={{HP-GAN:} Probabilistic 3D Human Motion Prediction via {GAN}},
  author={Barsoum, Emad and Kender, John and Liu, Zicheng},
  booktitle= CVPRW,
  pages={1418--1427},
  year={2018}
}

@inproceedings{kundu2019bihmp,
  author       = {Jogendra Nath Kundu and
                  Maharshi Gor and
                  R. Venkatesh Babu},
  title        = {BiHMP-GAN: Bidirectional 3D Human Motion Prediction {GAN}},
  booktitle    = AAAI,
  pages        = {8553--8560},
  year         = {2019},
}

@inproceedings{walker2017pose,
  title={The pose knows: Video forecasting by generating pose futures},
  author={Walker, Jacob and Marino, Kenneth and Gupta, Abhinav and Hebert, Martial},
  booktitle= ICCV,
  pages={3332--3341},
  year={2017}
}

@inproceedings{yan2018mt,
  title={Mt-vae: Learning motion transformations to generate multimodal human dynamics},
  author={Yan, Xinchen and Rastogi, Akash and Villegas, Ruben and Sunkavalli, Kalyan and Shechtman, Eli and Hadap, Sunil and Yumer, Ersin and Lee, Honglak},
  booktitle= ECCV,
  pages={265--281},
  year={2018}
}

@inproceedings{mao2021generating,
  title={Generating smooth pose sequences for diverse human motion prediction},
  author={Mao, Wei and Liu, Miaomiao and Salzmann, Mathieu},
  booktitle= ICCV,
  pages={13309--13318},
  year={2021}
}

@inproceedings{cai2021unified,
  title={A unified 3d human motion synthesis model via conditional variational auto-encoder},
  author={Cai, Yujun and Wang, Yiwei and Zhu, Yiheng and Cham, Tat-Jen and Cai, Jianfei and Yuan, Junsong and Liu, Jun and Zheng, Chuanxia and Yan, Sijie and Ding, Henghui and others},
  booktitle= ICCV,
  pages={11645--11655},
  year={2021}
}

@inproceedings{chen2023humanmac,
  title={Humanmac: Masked motion completion for human motion prediction},
  author={Chen, Ling-Hao and Zhang, Jiawei and Li, Yewen and Pang, Yiren and Xia, Xiaobo and Liu, Tongliang},
  booktitle= ICCV,
  pages={9544--9555},
  year={2023}
}

@inproceedings{wei2023human,
  title={Human joint kinematics diffusion-refinement for stochastic motion prediction},
  author={Wei, Dong and Sun, Huaijiang and Li, Bin and Lu, Jianfeng and Li, Weiqing and Sun, Xiaoning and Hu, Shengxiang},
  booktitle= AAAI,
  volume={37},
  pages={6110--6118},
  year={2023}
}

@article{yuan2019diverse,
  title={Diverse trajectory forecasting with determinantal point processes},
  author={Yuan, Ye and Kitani, Kris},
  journal={arXiv preprint arXiv:1907.04967},
  year={2019}
}

@inproceedings{gurumurthy2017deligan,
  title={Deligan: Generative adversarial networks for diverse and limited data},
  author={Gurumurthy, Swaminathan and Kiran Sarvadevabhatla, Ravi and Venkatesh Babu, R},
  booktitle= CVPR,
  pages={166--174},
  year={2017}
}

@article{dilokthanakul2016deep,
  title={Deep unsupervised clustering with gaussian mixture variational autoencoders},
  author={Dilokthanakul, Nat and Mediano, Pedro AM and Garnelo, Marta and Lee, Matthew CH and Salimbeni, Hugh and Arulkumaran, Kai and Shanahan, Murray},
  journal={arXiv preprint arXiv:1611.02648},
  year={2016}
}

@inproceedings{bhattacharyya2018accurate,
  title={Accurate and diverse sampling of sequences based on a “best of many” sample objective},
  author={Bhattacharyya, Apratim and Schiele, Bernt and Fritz, Mario},
  booktitle= CVPR,
  pages={8485--8493},
  year={2018}
}

@inproceedings{yuan2020dlow,
  title={Dlow: Diversifying latent flows for diverse human motion prediction},
  author={Yuan, Ye and Kitani, Kris},
  booktitle= ECCV,
  pages={346--364},
  year={2020}
}

@inproceedings{ma2022multi,
  title={Multi-objective diverse human motion prediction with knowledge distillation},
  author={Ma, Hengbo and Li, Jiachen and Hosseini, Ramtin and Tomizuka, Masayoshi and Choi, Chiho},
  booktitle= CVPR,
  pages={8161--8171},
  year={2022}
}

@inproceedings{salzmann2022motron,
  title={Motron: Multimodal probabilistic human motion forecasting},
  author={Salzmann, Tim and Pavone, Marco and Ryll, Markus},
  booktitle= CVPR,
  pages={6457--6466},
  year={2022}
}

@inproceedings{dang2022diverse,
  title={Diverse human motion prediction via gumbel-softmax sampling from an auxiliary space},
  author={Dang, Lingwei and Nie, Yongwei and Long, Chengjiang and Zhang, Qing and Li, Guiqing},
  booktitle= ACMMM,
  pages={5162--5171},
  year={2022}
}

@inproceedings{mdm,
  title={Human motion diffusion model},
  author={Tevet, Guy and Raab, Sigal and Gordon, Brian and Shafir, Yonatan and Cohen-Or, Daniel and Bermano, Amit H},
  booktitle= ICLR,
  year={2023}
}

@inproceedings{motiongpt,
  title={Motiongpt: Human motion as a foreign language},
  author={Jiang, Biao and Chen, Xin and Liu, Wen and Yu, Jingyi and Yu, Gang and Chen, Tao},
  booktitle= NeurIPS,
  pages={20067--20079},
  year={2023}
}

@inproceedings{motionlm,
  title={Motionlm: Multi-agent motion forecasting as language modeling},
  author={Seff, Ari and Cera, Brian and Chen, Dian and Ng, Mason and Zhou, Aurick and Nayakanti, Nigamaa and Refaat, Khaled S and Al-Rfou, Rami and Sapp, Benjamin},
  booktitle= ICCV,
  pages={8579--8590},
  year={2023}
}

@inproceedings{curreli2025nonisotropic,
  title={Nonisotropic Gaussian Diffusion for Realistic 3D Human Motion Prediction},
  author={Curreli, Cecilia and Muhle, Dominik and Saroha, Abhishek and Ye, Zhenzhang and Marin, Riccardo and Cremers, Daniel},
  booktitle= CVPR,
  pages={1871--1882},
  year={2025}
}

@STRING{PAMI  = "IEEE Transactions on Pattern Analysis and Machine Intelligence"}

@STRING{IJCV  = "International Journal of Computer Vision"}

@STRING{CVPR     = "Proceedings of the IEEE/CVF Conference on Computer Vision and Pattern Recognition"}

@STRING{CVPRW     = "Proceedings of the IEEE/CVF Conference on Computer Vision and Pattern Recognition Workshops"}

@STRING{ICCV     = "Proceedings of the IEEE/CVF International Conference on Computer Vision"}

@STRING{ECCV     = "Proceedings of the European Conference on Computer Vision"}

@STRING{NeurIPS  = "Advances in Neural Information Processing Systems"}

@STRING{ICML     = "Proceedings of the International Conference on Machine Learning"}

@STRING{ICLR     = "Proceedings of the International Conference on Learning Representations"}

@STRING{AAAI     = "Proceedings of the AAAI Conference on Artificial Intelligence"}

@STRING{IJCAI    = "Proceedings of the International Joint Conference on Artificial Intelligence"}

@STRING{WACV     = "Proceedings of the IEEE/CVF Winter Conference on Applications of Computer Vision"}

@STRING{BMVC     = "Proceedings of the British Machine Vision Conference"}

@STRING{ICRA     = "Proceedings of the IEEE International Conference on Robotics and Automation"}

@STRING{IROS     = "Proceedings of the IEEE/RSJ International Conference on Intelligent Robots and Systems"}

@STRING{ACMMM    = "Proceedings of the ACM International Conference on Multimedia"}
\bibliographystyle{tmlr}

\appendix
\begin{figure}[t]
    \centering
    \includegraphics[width=\linewidth]{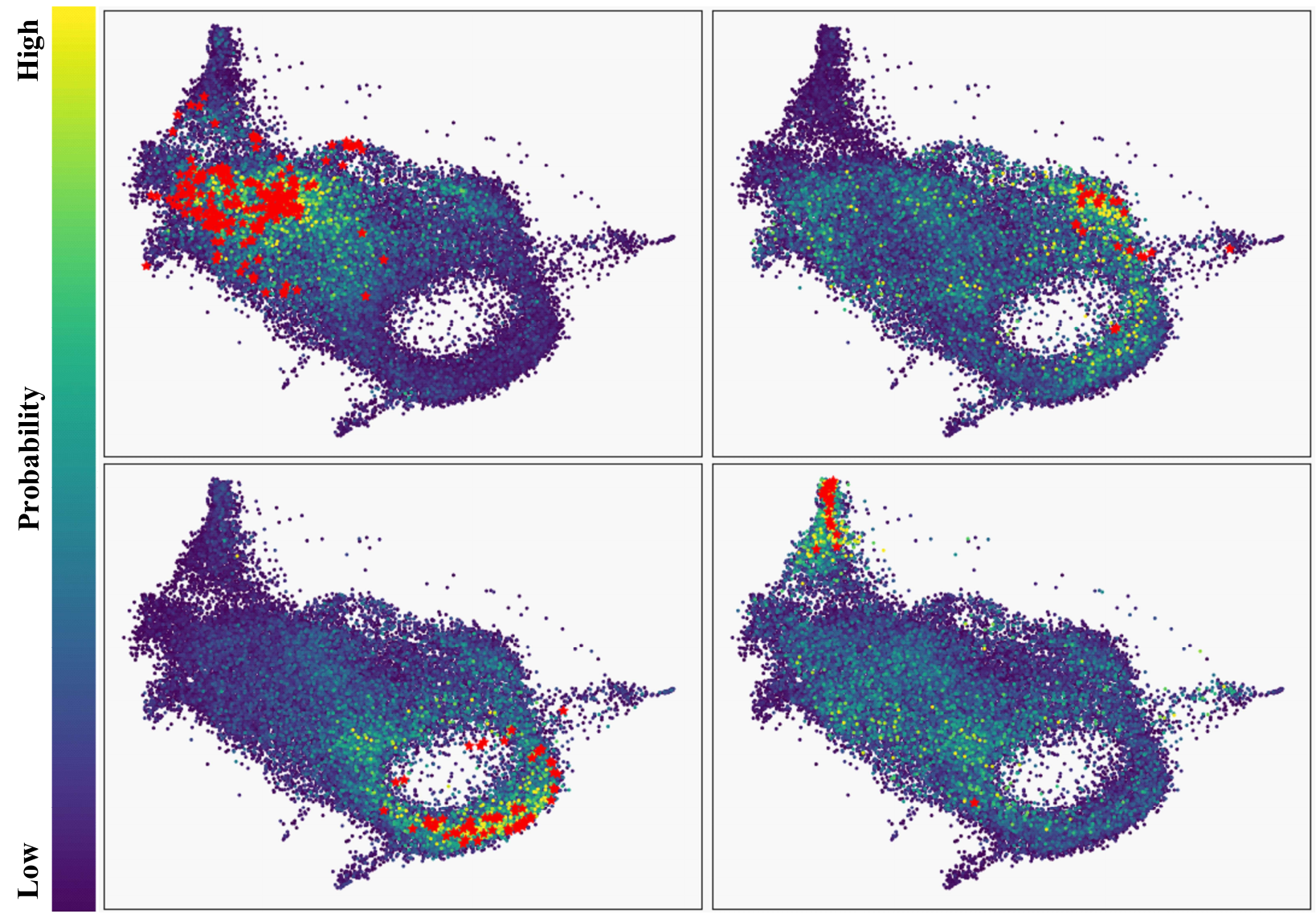}
    \caption{\textbf{Visualization of future motion densities by \methodname{}.} The estimated densities for four different motion sequences are visualized. We used UMAP to project these future motions onto a 2D space. Each dot represents an evaluated future motion, and the color of each dot indicates its probability, as shown in the side color bar. The red stars represent the projected \rev{multimodal ground truth future motions~\citep{yuan2019diverse}. Multimodal ground truths are defined by grouping similar motions, which can be interpreted as multiple distinct possibilities.} }
    \label{fig:density}
\end{figure}

\section{Reproducibility Statement}

We are committed to ensuring the reproducibility of our results. To this end, we will release the full source code, including training and evaluation scripts. All datasets used in this study (Human3.6M \citep{ionescu2013h36m}, AMASS \citep{mahmood2019amass}, and derived benchmarks) are publicly available. Our implementation is based on PyTorch, and all experiments except \cref{subsubsec:cost} were conducted on a workstation with an Intel Xeon Gold 6154 CPU and an NVIDIA A100 GPU. 

We report detailed computational and memory costs in Table~\ref{tab:cost}, and we describe architectural and training hyperparameters in Appendix~B. For reproducibility, we specify the number of training epochs, optimizer (Adam with $\beta_1{=}0.9$, $\beta_2{=}0.999$, learning rate $5 \times 10^{-4}$), batch sizes (192 for Human3.6M, 64 for AMASS). Random seeds were fixed to 0 across all experiments.

\section{Implementation Details of Kernel Density Estimation}

We assessed the accuracy of density estimation using Kernel Density Estimation (KDE) on previous methods. To ensure a fair comparison of inference time, all KDE computations were conducted on the GPU. We applied KDE to the standardized predicted future motions (or latents for BeLFusion) to obtain the estimated density. 
In this process, the $i$-th dimension of the predicted future motions was standardized using its $i$-th variance, meaning that covariances were not considered during standardization. 
We employed Scott's rule to determine the optimal bandwidth for KDE.

\section{Visualization of Estimated Density}
\label{sec:vis_estimated_density}

We visualized the future motion density estimated by \methodname{}. Since future motions are high-dimensional data, we used UMAP~\citep{DBLP:journals/corr/abs-1802-03426} to project each future motion into a 2D space. We displayed the multimodal ground truth future motions~\citep{yuan2019diverse} alongside the visualized density map. \rev{Multimodal ground truths are defined by grouping similar motions, which can be interpreted as multiple distinct possibilities.} As shown in ~\cref{fig:density}, \methodname{} estimated a high probability around the ground truth in all motion sequences. This visually supports the high density estimation accuracy presented in \cref{tab:probability}.

\section{Comparison to GMM Only Method}
While we propose a combination of a conditional Gaussian mixture model and unconditional flow matching, a Gaussian Mixture Model (GMM)-only baseline might appear sufficient to model the conditional distribution. However, our experiments reveal that this is not the case. As shown in \cref{tab:GMM}, this approach significantly degrades performance across all metrics. While a GMM itself can approximate conditional distributions, it lacks the expressive power to capture complex latent distributions effectively. 

\begin{table}[t]
\caption{Comparison to the GMM only model (without Flow Matching) on Human3.6M.}
\centering
\begin{tabular}{l|lllll}
 & APD & ADE & FDE & MMADE & MMFDE \\ \hline
\methodname{} & 6.097 & 0.369 & 0.472 & 0.481 & 0.510 \\
GMM only & 4.545 & 0.548 & 0.751 & 0.637 & 0.763
\end{tabular}
\label{tab:GMM}
\end{table}

\section{Performance Gap on APD}
While APD measures diversity via pairwise distances between generated samples, higher APD does not always imply better diversity. In fact, excessively high APD can indicate unrealistic or noisy predictions that ignore the input context.

For example, GSPS~\citep{mao2021generating} and DivSamp~\citep{dang2022diverse} report very high APDs (14.757 and 15.310), but we observed failure cases such as unrealistic floating motions, as also mentioned in other paper~\citep{chen2023humanmac}. In contrast, recent diffusion-based models (BeLFusion~\citep{barquero2023belfusion}, HumanMAC~\citep{chen2023humanmac}, CoMusion~\citep{sun2024comusion}) prioritize plausibility and contextual consistency, resulting in moderate APDs (7.602, 6.301, 7.632).

Our method achieves a comparable APD (6.101) to these diffusion-based methods, indicating a good balance between diversity and realism. Furthermore, we evaluated Frechet Inception Distance (FID), which better captures semantic and distributional similarity to real data. As shown in ~\cref{tab:apd}, \methodname{} achieves significantly better FID than GSPS and DivSamp, supporting our claim that our method generates realistic and diverse future motions.

\begin{table}[t]
\caption{The APD (Average Pairwise Distance) and FID (Frechet Inception Distance) on the previous methods and \methodname{} on Human3.6M.}
\centering
\begin{tabular}{l|ll}
 & APD & FID \\ \hline
GSPS & 14.757 & 2.103 \\
DivSamp & 15.310 & 2.083 \\
BeLFusion & 7.602 & 0.209 \\
HumanMAC & 6.301 & - \\
CoMusion & 7.632 & 0.102 \\
\methodname{}(Ours) & 6.101 & 0.196
\end{tabular}
\label{tab:apd}
\end{table}

\section{Cache Augmentation and Adaptation}
\methodname{} is naturally suited for augmentation and adaptation. New motions can be added incrementally without retraining, simply by recomputing cached flows for the additional data. When transferring to a new dataset, cache recomputation provides fast adaptation and requires significantly fewer resources than retraining the full flow. Furthermore, since the cache acts as an external memory, continual learning becomes feasible without catastrophic forgetting. This property makes \methodname{} practical for long-term deployment in real-world scenarios such as assistive robotics and animation pipelines.

\section{Potential Broader Impact}
\label{sec:broader_impact}

Research in human motion prediction, including our proposed \methodname{}, carries significant societal implications that must be carefully considered. Our primary goal is to advance computational understanding of human movement, enabling positive applications in fields like biomechanics, physical rehabilitation, robotics, and immersive virtual reality environments. However, we acknowledge the potential for misuse and the ethical challenges inherent in this technology.

\subsection{Potential Positive Impacts}

The proposed \methodname{} introduces a fast probaility-aware motion prediction framework,
which may involve the following broader impacts:

\begin{itemize}
    \item \textbf{Improved Collaboration in Robotics and Automation.} In collaborative robotics and industrial automation, understanding and anticipating human motion is critical for ensuring safety and efficiency. The proposed system enables robots to predict human actions and movements with probabilistic confidence, allowing them to adjust their trajectories and tasks in real time. This leads to smoother coordination in shared workspaces such as manufacturing floors, warehouses, or hospitals, where humans and robots must work in close proximity.
    \item \textbf{Proactive Support in Assistive Technologies.} In assistive technologies for the elderly and individuals with disabilities, anticipating human motion is essential for delivering timely and meaningful support. A fast and uncertainty-aware human motion prediction system enables robots and smart devices to proactively assist users by foreseeing movements such as standing, walking, or reaching, even in the presence of noisy or partial sensor data. Furthermore, such a system could help prevent falls or injuries by detecting signs of instability and initiating interventions early.
    \item \textbf{Immersive Interactions in VR and Gaming.} Virtual reality (VR) and gaming systems stand to benefit from predictive models that can estimate future body movements in real time with associated uncertainties. This capability allows VR applications to reduce latency and create more responsive environments by anticipating user actions and gestures.
\end{itemize}

\subsection{\rev{Ethical Concerns and Mitigation Strategies}}

We recognize two main areas of ethical concern: privacy and potential misuse.

\subsubsection{Privacy and Surveillance}
Systems that predict human motion are often trained on public datasets (like AMASS and 3DPW) that contain recordings of real individuals. While these datasets are typically anonymized by representing the subject as a skeletal mesh (e.g., SMPL model) without texture or identity information, the technology itself could be integrated into real-time surveillance systems. The ability to predict future movement can enhance tracking and identification. As researchers, we strongly oppose the use of this technology for unauthorized surveillance or discriminatory profiling. Our research focus is strictly on foundational scientific advancement and is confined to lab-controlled data and applications that respect user consent.

\subsubsection{Bias and Fairness in Prediction}
Bias in motion datasets is a recognized issue, as they may not equally represent all human populations across factors such as body shape, age, gender, culture, or ability. If \methodname{} were deployed on biased data, its predictive accuracy could be lower for underrepresented groups, potentially leading to critical failures in safety-critical applications (e.g., in advanced driver-assistance systems or elderly care robotics). We mitigate this by using large, diverse datasets (like AMASS) and demonstrating robust generalization to out-of-distribution data (3DPW), but we emphasize that continued dataset development focusing on fairness and inclusivity is critical for the field.

\subsubsection{Dual-Use Concerns}
The core technological capability of anticipating human actions presents a dual-use challenge. While the positive applications are numerous, the technology could be leveraged to anticipate behavior in adversarial or unauthorized contexts. Our work is purely academic and aims to contribute to the open body of knowledge for socially beneficial purposes. We encourage the responsible deployment of our method only in contexts with clear consent and explicit ethical guidelines.

\end{document}